\documentclass{article}

\usepackage{microtype}
\usepackage{graphicx}
\usepackage{subfigure}
\usepackage{booktabs} % for professional tables
\usepackage{bm}
\usepackage{amsmath,amssymb,amsfonts}
\usepackage{comment,todonotes}
\usepackage{nicefrac}
\usepackage{dblfloatfix}    % To enable figures at the bottom of page
\usepackage{placeins}

\usepackage[symbol]{footmisc}

\usepackage{hyperref}
\newcommand{\name}{VI-HDS}

\newcommand{\modelWB}{constant_dev6_s0}
\newcommand{\modelBB}{BB_X5_Y2_Z5_S1_s0}

\newcommand{\popcolor}{blue}
\newcommand{\groupcolor}{magenta}
\newcommand{\indcolor}{black}

\newcommand{\noisestate}{\mathbf{v}}

\newcommand{\neuralfunction}{\omega}
\newcommand{\nnargs}{ \mathbf{\Psi}}

\newcommand{\vx}{\mathbf{x}}
\newcommand{\vw}{\mathbf{w}}

\newcommand{\mx}{\mathbf{X}}
\newcommand{\mobs}{\mathbf{Y}}

\newcommand{\muY}{\mathbf{M}}
\newcommand{\varY}{\mathbf{\Sigma}}

\newcommand{\nspecies}{M}

\newcommand{\ntime}{T}

\newcommand{\observerprocess}{\psi}
\newcommand{\noiseprocess}{\rho}

\newcommand{\vgroup}{\mathbf{g}}

\newcommand{\popvaridx}{l}
\newcommand{\groupvaridx}{j}
\newcommand{\indvaridx}{k}

\newcommand{\groupvarid}{G}
\newcommand{\popvarid}{P}
\newcommand{\indvarid}{I}

\newcommand{\groupvar}{G}
\newcommand{\popvar}{P}
\newcommand{\indvar}{I}

\newcommand{\vcassette}{\mathbf{c}}

\newcommand{\vtreat}{\mathbf{u}}

\newcommand{\vy}{\mathbf{y}}
\newcommand{\vz}{\mathbf{z}}
\newcommand{\vtheta}{\bm{\theta}}
\newcommand{\vphi}{\bm{\phi}}

\newcommand{\ndevicesub}{K}

\newcommand{\nsubs}{S}
\newcommand{\dataidx}{n}

\newcommand{\devicesub}{s}

\newcommand{\vu}{\mathbf{u}}

\newcommand{\vo}{\mathbf{o}}

\usepackage[accepted]{icml2019}

\icmltitlerunning{Efficient Amortised Bayesian Inference for Hierarchical and Nonlinear Dynamical Systems}

\begin{document}

\twocolumn[
\icmltitle{Efficient Amortised Bayesian Inference \\ for Hierarchical and Nonlinear Dynamical Systems}
%\icmltitle{Variational Inference for Neural Mixed-Effects Models of Biological Systems}
%\icmltitle{Neural Mixed-Effects Models of Synthetic Biological Circuits}
% Stochastic Variational Inference for Nonlinear Mixed-Effects ODEs}



\icmlsetsymbol{equal}{*}

\begin{icmlauthorlist}
\icmlauthor{Geoffrey Roeder}{micro,pton}
\icmlauthor{Paul K Grant}{micro}
\icmlauthor{Andrew Phillips}{micro}
\icmlauthor{Neil Dalchau}{micro}
\icmlauthor{Edward Meeds}{micro}
\end{icmlauthorlist}

\icmlaffiliation{micro}{Microsoft Research, Cambridge, United Kingdom}
\icmlaffiliation{pton}{Princeton University, Princeton, United States of America}

\icmlcorrespondingauthor{Geoffrey Roeder}{roeder@princeton.edu}
\icmlcorrespondingauthor{Ted Meeds}{edmeeds@microsoft.com}
\icmlcorrespondingauthor{Neil Dalchau}{ndalchau@microsoft.com}

\icmlkeywords{Machine Learning, Deep Learning, Bayesian Statistics, Synthetic Biology, ICML 2019}
\vskip 0.3in
]



\printAffiliationsAndNotice{}  % leave blank if no need to mention equal contribution

\begin{abstract}
We introduce a flexible, scalable Bayesian inference framework for nonlinear dynamical systems characterised by distinct and hierarchical variability at the individual, group, and population levels. Our model class is a generalisation of nonlinear mixed-effects (NLME) dynamical systems, the statistical workhorse for many experimental sciences.
We cast parameter inference as stochastic optimisation of an end-to-end differentiable, block-conditional variational autoencoder. We specify the dynamics of the data-generating process as an ordinary differential equation (ODE) such that both the ODE and its solver are fully differentiable.
This model class is highly flexible: the ODE right-hand sides can be a mixture of user-prescribed or ``white-box" sub-components and neural network or ``black-box" sub-components.  Using stochastic optimisation, our amortised inference algorithm could seamlessly scale up to massive data collection pipelines (common in labs with robotic automation).  Finally, our framework supports interpretability with respect to the underlying dynamics, as well as predictive generalization to unseen combinations of group components (also called ``zero-shot" learning).   
We empirically validate our method by predicting the dynamic behaviour of bacteria that were genetically engineered to function as biosensors.
\end{abstract}

\section{Introduction}

Dynamical systems have been the central focus in many areas of statistical machine learning, whether as state-space models \cite{archer2015black, karl2016deep}, (non)linear dynamical systems and their extensions \cite{becker2018switching, johnson2016structured, linderman2016recurrent}, or neuroscience applications \cite{wu2017gaussian, gao2016linear}.
Recent work has even reinterpreted general recursive deep neural networks as discretisations of continuous-time dynamical systems \cite{chen2018neural}.

Dynamical systems learned from experimental data are widespread in the physical sciences, including fluid dynamics, thermodynamics, and electromagnetism.
They also play a particularly important role in advancing our understanding of biology, typically studied as Ordinary Differential Equations (ODEs).
Seminal contributions of ODE models in biology include the mechanisms of the cell cycle \citep{Tyson2001,Ferrell2011}, circadian rhythms \citep{leloup2003toward}, and nerve action potentials \citep{Hodgkin1952}. 

More recently, ODE models have been playing an increasingly important role in engineering biological systems by capturing mechanistic processes in the underlying chemical reaction dynamics that identify potential targets for molecular intervention and by predicting the possible outcomes of these interventions \citep{Elowitz2000,Nathan2016,Gardner2000}.
The ability to precisely engineer biology could enable substantial breakthroughs in medicine and crop yields, and provide sustainable alternatives to environmentally unsustainable processes and products \citep{Khalil2010, Ruder2011, Cumbers2017}.
However, identifying the parameters of ODE models is still limited by the efficiency and scalability of inference methods \cite{calderhead2011statistical}. 
Markov Chain Monte Carlo (MCMC) is considered state-of-the-art for learning ODE parameters from experimental data \cite{woods2016statistical, moore2018rapid}, but requires expensive numerical integration at each time step. 

This problem is exacerbated by  model selection, where competing explanatory mechanisms must be evaluated for predictive accuracy and generalisation \citep{calderhead2009estimating}.
In the era of increasingly automated laboratories, scientists require new statistical models and computational methods that can scale adequately with the increasing availability of experimental data.

To address these computational and modeling challenges, we introduce {\bf \name}: \textbf{V}ariational \textbf{I}nference over the parameters of \textbf{H}ierarchical \textbf{D}ynamical \textbf{S}ystems that have a defined generative process and likelihood.\footnote[1]{Our implementation of VI-HDS, the dataset, and all code to reproduce the experimental results is available at \url{https://www.github.com/Microsoft/vi-hds}.}
Specifically, we cast  inference over the parameters of a dynamical system with hierarchical or multi-level structure as optimisation of a conditional variational autoencoder \cite{sohn2015learning, kingma2013auto, rezende2014stochastic}.

We interpret this framework as a generalisation of ``nonlinear mixed-effects models" (NMEMs) \citep{ davidian2003nonlinear}---a.k.a. hierarchical or multilevel nonlinear models---which are workhorse statistical models in the experimental sciences \cite{fitzmaurice2008longitudinal}.
Examples of modelling problems suitable for NMEMs includes a population of human cells with different combinations of shared proteins, or a population of engineered bacterial cells with different combinations of shared genetic sequences. 
Through independence assumptions in the probabilistic model, NMEMs separate inference over individual, group, and population-level variability \cite{karlsson2015nonlinear}.
In \name, we design block conditioning in the variational distribution to achieve the same hierarchical structure for an arbitrary dynamical system, improving the model fit and predictive accuracy.

We validate our proposed framework with a case study in synthetic biology: modelling the activity of synthetic biological devices in bacteria.
We show that \name{} has the following desirable properties:
%\begin{itemize}
(i) performs as well as state-of-the-art methods but is substantially faster to learn, while enabling 
(ii) scaling to massive datasets by leveraging fast reverse-mode automatic differentiation and amortised inference,
(iii) easily incorporates related dynamical systems for efficient learning by sharing statistical strength,
(iv) facilitates zero-shot learning through compositional group structure;
(v) enables rapid closed-loop iteration across plausible mechanistic models, which can be a mixture of prescribed equations and black-box neural networks.
\section{Methodology}

\begin{figure*}[ht]
	\begin{tabular}{p{11.5cm}p{4.5cm}}
		{\sf A} & {\sf B} \\
		\includegraphics[trim={0 120 0 200}, clip,width=\linewidth]{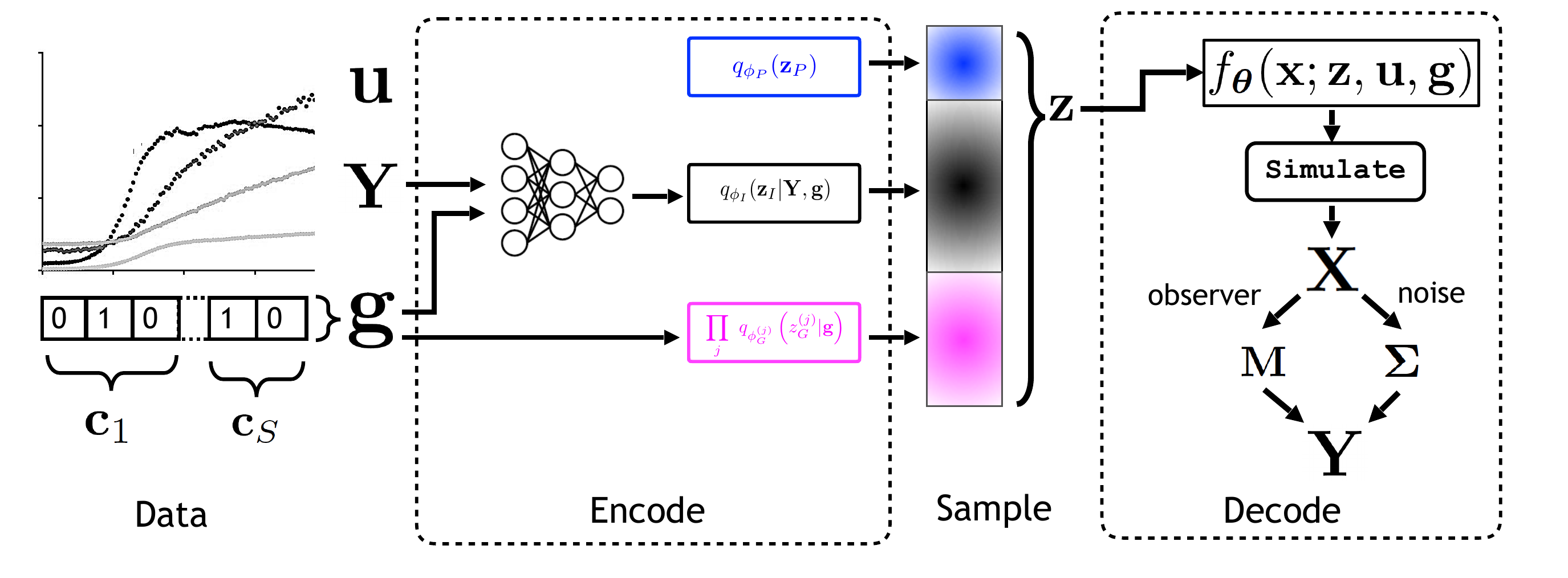}
		&
		\includegraphics[trim={0 20 0 0}, clip, width=\linewidth]{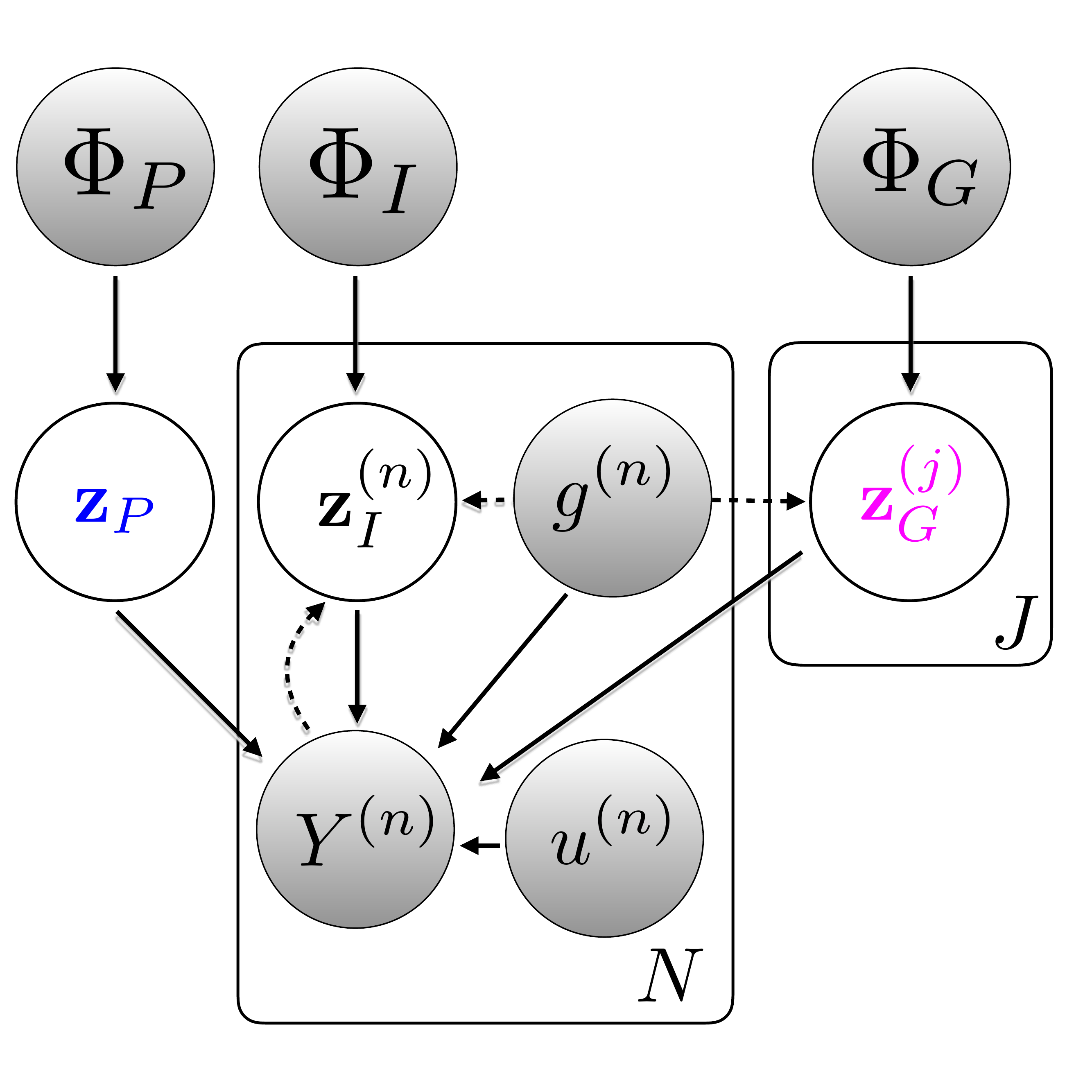}
	\end{tabular}
\vspace{-0.1in}
\caption{{\bf Auto-encoding hierarchical and nonlinear dynamical systems}. {\bf (A)}: The computational flow graph for encoding, sampling from the variational posterior, and simulating the dynamical system. Note that the sample and simulate operations are constrained to be differentiable. {\bf (B)}: Probabilistic structure for the sequence-valued dependent variable $\mobs$. Dashed lines represent dependencies in $q$, and solid lines in $p$. $\Phi$ represents prior information elicited from a domain expert.
}
\label{fig:encoder_schematic}
\end{figure*}

In this section we describe our hierarchical and nonlinear dynamical systems model, followed by posterior inference framework using conditional variational autoencoders.
We present detailed descriptions of the observable data, followed by the generative process and the variational inference approach.  
We then discuss how to learn a multilevel generative model alongside a conditional variational posterior.
Finally, we demonstrate the flexibility of our approach through the specification of a dynamical system by either prescribed mechanistic equations ({\em white-box}) or deep neural networks ({\em black-box}).  Our probabilistic assumptions are summarised by the graphical model in Figure~\ref{fig:encoder_schematic}B.

\subsection{Data Model}
\label{sec:data_model}
\paragraph{Repeated observations of groups}
In many natural sciences it is common to refer to both the observation variables and the state variables as {\em species}, because they directly or indirectly represent a {\em kind} or {\em type} of measurable quantity in the real world, such as protein concentrations.
We adopt this convention here.
The $\dataidx$th datum $\mobs^{(n)}$ is an $\nspecies \times \ntime$ matrix, where $\nspecies$ is the number of {\em observed species}, repeatedly measured over $\ntime$ time points. 
In order to support variability in closely related observations, we introduce the random variable $\vtreat$ to represent such experimental conditions.

\paragraph{Group stratification by composition}
We model a population of dynamical systems, each containing a different combination of $\nsubs$ shared components.
A specific combination of components defines a \emph{group}, which we formalise as a multi-hot vector $\vgroup = \left[ \vcassette^{\top}_1, \ldots, \vcassette^{\top}_{\nsubs} \right]^{\top}$, where  
each  $\vcassette_{\devicesub}$ is a one-hot vector that identifies $1$ of $\ndevicesub^{(\devicesub)}$ components.
The multi-hot vector $\vgroup$ has exactly $S$ non-zero entries. 
By composing groups this way, we can perform \emph{zero-shot learning}, where our model learns to predict the behaviour of unseen groups, as described in Section~\ref{sec:blockencoder} and studied experimentally in Section~\ref{sec:held_out}.

We now turn to a description of the generative process, which motivates our conditional variational autoencoder inference algorithm.

\subsection{Generative Process}
\label{sec:mixed_effect_meaning}
We model observations using a latent variable model with the generative process described in Equations~\ref{eq:prior}-\ref{eq:likelihood}.  Latents $\vz$ represent  both parameters of the dynamical systems model, and, if necessary, noise process parameters (e.g. variances in the likelihood).  We emphasise that the structure of $p_{\vtheta}(\vz | \vgroup)$ will depend on whether the model is a white-box, whereby the prior will contain informative domain knowledge, or a black-box, where it will be uninformative diagonal Gaussians.
\begin{align}
\vz &\sim p_{\vtheta}(\vz | \vgroup) \label{eq:prior} \\
\dot{\vx} &= f_{\vtheta}(\vx;\vz,\vtreat,\vgroup) \label{eq:ode} \\
\mx &= \texttt{Simulate}( f_{\vtheta}, \vx_0 ) \label{eq:simulate} \\
\muY &= \observerprocess(\mx ), ~~~~ \varY = \noiseprocess( \mx, \vz ) \label{eq:observer_and_noise}\\
\mobs &\sim p( \mobs | \muY, \varY ) \label{eq:likelihood}
\end{align}
Central to the generative process is $f_{\vtheta}$, the dynamics model describing the  behaviour of an individual's unobserved state vector $\vx$ over time, where $\dot{\vx}$ denotes the time derivative of $\vx$. 
Along with latent variables $\vz$ and group indicators $\vgroup$, $f$ depends on $\vtreat$, a vector of observed \emph{experimental conditions} or {\em treatments}.  
Solving the dynamical system results in a state-time matrix $\mx$.  
The state-time matrix is then mapped to mean $\muY$ and, along with $\vz$, covariance $\varY$  of $\mobs$ through an observation process $\observerprocess$ and noise process $\noiseprocess$. 

For generality, in \eqref{eq:observer_and_noise}, we have included observer  $\observerprocess$ and noise process $\noiseprocess$ functions.
For dynamical systems that don't require such processes, $\observerprocess$ can be the identity function and $\noiseprocess$ some fixed covariance matrix.

The dynamics model $f$ has parameters $\vtheta$ that are adjusted during learning: for the white-box, they are group-level offsets and scales applied to $\vz_{\groupvar} \in \vz$; for the black-box they are neural network weights.  
In all our experiments we assume $p( \mobs | \muY, \varY )$ to be Gaussian.
The exact form of the noise process $\noiseprocess$ depends on which model we use, and will be explained below according to the white- and black-box framework. 

Note that we have written \eqref{eq:ode} and \eqref{eq:simulate} as a general dynamical system.   For processes that exhibit time-varying noise, we can modify \eqref{eq:ode} and \eqref{eq:simulate} with a time-dependent noise process, to represent the diffusion and drift processes of a stochastic differential equation.  
However, for clarity of exposition, we will focus on ODE systems as the representative dynamical system in the remainder. 

\subsection{Auto-Encoding Hierarchical and Nonlinear Dynamical Systems}
Core to our approach is amortised stochastic variational inference \cite{hoffman2013stochastic} with a reparameterised variational distribution  \cite{kingma2013auto, rezende2014stochastic}, a.k.a. a variational auto-encoder (VAE). 
We extend this basic framework to hierarchical structure in the dynamical system through conditional variational autoencoders \citep{sohn2015learning}.
We assume a conditional prior $p_{\vtheta}(\vz | \vgroup)$, conditional generative $p_{\vtheta}( \mobs | \vz, \vgroup, \vtreat )$, and conditional variational $q_{\vphi}( \vz | \mobs, \vgroup, \vtreat )$ distributions whose parameters are directly or indirectly calculated using neural networks.  Figure~\ref{fig:encoder_schematic}A illustrates the computational flow of our framework.

Neural networks weights $\vphi$ and $\vtheta$ are adjusted by stochastic gradient ascent to maximise the evidence lower bound (ELBO) on the conditional log marginal data likelihood  $\log ~p(\mobs | \vgroup, \vtreat)$:
\begin{align}
\mathbb{E}_{q_{\vphi}( \vz | \mobs, \vgroup, \vtreat )}[  \log p_{\vtheta}( \mobs | \vz, \vgroup, \vtreat ) &+ \log p_{\vtheta}(\vz | \vgroup)  \nonumber \\
&- \log q_{\vphi}( \vz | \mobs, \vgroup ) ] \label{eq:vaeelbo}
\end{align}
VAE training thus simultaneously learns an amortised inference model $q$ and a generative model $p$, transforming traditional inference over an intractable integral into a gradient-based optimisation procedure, bringing significant computational and scaling benefits to learning and inference.   
We note that this inference method can also be applied to ``simple" mechanistic models (i.e. a fixed decoder model) using likelihood-free inference \cite{moreno2016automatic, tran2017hierarchical}.

\subsection{Multilevel Block-Conditional Encoder}\label{sec:blockencoder}
Our approximate posterior $q_{\vphi} \left(\vz | \mobs, \vgroup \right)$ is factorised into three blocks of independent latent variables representing population, group, and individual factors: 
\begin{equation*}
\textcolor{\popcolor}{\underbrace{ ~q_{\phi_{\popvarid}} \left( \vz_{\popvar} \right)}_\text{Population}}~
\textcolor{\indcolor}{\underbrace{ ~q_{\phi_{\indvarid}} \left( \vz_{\indvar}| \mobs,  \vgroup \right)}_\text{Individual}}
\textcolor{\groupcolor}{\underbrace{ ~\prod_{\groupvaridx}~
		q_{\phi_{\groupvarid}^{(\groupvaridx)}} \left( \vz_{\groupvar}^{(\groupvaridx)}| \vgroup \right)}_\text{Group}} ~~~
\end{equation*}
This three-level hierarchy allows us to match latent variable capacity to individual, group, and population level variability according to our model of the data generating process.
For example, if the majority of the variability occurs at the group level, more capacity in the variational posterior can be assigned to the group-level block through additional dimensions in $\vz_{\groupvarid}$.  Note this flexibility generally applies only to the black-box model, since the white-box parameter hierarchy is prescribed by domain knowledge.  
Next, we formalise the notion of encoding variability at multiple hierarchical levels.
Note that we make a mean-field assumption for the variational posterior, but in general, any distributions may be chosen.

{\bf Population-level latent variables} are independent of observations and groups. For each of the L dimensions of the population block of the mean-field posterior, we set
% switch to sigma^2 and multiplied by 2 in log
\begin{align*}
{\color{\popcolor} q_{\vphi_{\popvarid}(\popvaridx)} = \mathcal{N} \left( \mu = r_{\popvaridx}, \sigma^2 = \exp \{2 v_{\popvaridx}\} \right)},
\end{align*}
where {\color{\popcolor} $ \vphi_{\popvarid}(\popvaridx) = \{ r_{\popvaridx},v_{\popvaridx} \}$}, and
${\color{\popcolor} r_{\popvaridx}, v_{\popvaridx}} \in \mathbb{R}$.

{\bf Group-level latent variables} are independent of $\mobs$ but dependent on group membership. Recall that a group is a collection of components. 
As described in section \ref{sec:data_model}, we use the multi-hot {\color{\groupcolor}$\vgroup_\groupvaridx$} to form a summation of group-specific scalars, each of which corresponds to a component in the following way:
\begin{align*}
{\color{\groupcolor} q_{\vphi_{\groupvarid}^{(\groupvaridx)}(m)}  = \mathcal{N} \left( \mu =  \bm{\nu}_{m}^\top \vgroup_j,~\sigma^2 = \exp \{ 2 \bm{\eta}_{m}^\top \vgroup_j \} \right)},
\end{align*} where {\color{\groupcolor} $\vphi_{\groupvarid}(m) = \{ \bm{\nu}_{m}, \bm{\eta}_{m}\}$}, sharing parameters among the groups
through the selector mechanism {\color{\groupcolor}$\vgroup_{\groupvaridx}$}.

{\bf Individual-level latent variables} are encoded in a manner similar to typical VAEs.  
The weights and biases of a deep neural network are contained in $\vphi_I$. The networks encode the mean ${m}_k$ and variance ${ v}_k$ functions for the $k$'th factor, so that
\begin{align*}
q_{\vphi_{{\indvarid}}(\indvaridx)}= \mathcal{N} \left( \mu = {m}_{\indvaridx}(\mobs, \vgroup), \sigma^2 = { v}_{\indvaridx} (\mobs, \vgroup) \right).
\end{align*}
The encoder neural networks perform amortised inference that is optimised during learning.

Alternatively, individual parameters can be optimised per data instance, giving tighter ELBOs. However, non-amortised stochastic variational inference \cite{hoffman2013stochastic} would not scale with massive amounts of data---one of the key desiderata for our method.

\subsection{Block-Conditional Decoder}
\label{sec:decoder}
The bulk of the multilevel modelling occurs in the variational posterior, whose draws are used as parameters in the ODE model $f$. 
However, group-conditional modifications are made to $\vz_{\groupvar}$ in both the white-box and black-box decoder in order to induce the correct conditional probabilistic structure for the conditional ELBO. 
For example, before passing $\vz_{\groupvar}^{(\groupvaridx)}$ into $f$, we apply weights $\vw \in \vtheta$, modifying $\vz_{\groupvar}^{(\groupvaridx)} \leftarrow \vz_{\groupvar}^{(\groupvaridx)} * \exp(w_1^T \vgroup) + w_2^T \vgroup$, corresponding to a group-dependent Gaussian; this implies the correct conditional VAE for group-conditional variables\footnote{ Note that in many cases, variables are constrained to be positive, which we model as log-normals.  In this case, $\vz_{\groupvar}^{(\groupvaridx)}$ is implicitly a group-dependent log-normal.}.
The specific form of the block-conditional decoder depends on white- or black-box right-hand side choices for the dynamical system.
We discuss these choices next and show how to induce the appropriate multilevel probabilistic structure in each. 

\paragraph{Black-Box Dynamics.}
To simplify notation, let $\nnargs = \{ {\color{\popcolor} \vz_{\popvar}}, {\color{\groupcolor} \vz_{\groupvar}}, {\color{\indcolor} \vz_{\indvar}},  \vtreat, \vgroup \}$ and a function based on a neural network be $\neuralfunction$. 
We can write an ODE equation as $\dot{\vx} = \neuralfunction( \vx, \nnargs; \vtheta )$, where the dot indicates the first time derivative. 
In many cases, including our case study, prior domain knowledge regarding the functional form of the ODE can be incorporated into the black-box. 
For example, the following ODE is the difference between a growth and degradation term for the latent process, but we know that degradation will go to 0 when $\vx \rightarrow 0$. 
To encode this we use a softplus activation over two networks, denoted by $+$:
\begin{equation}
\dot{\vx} = \neuralfunction_1^{+}( \vx, \nnargs ) - \vx \odot \neuralfunction_2^{+}( \vx, \nnargs ) 
\label{eq:nnstate}
\end{equation}
where $\odot$ is the Hadamard product.

\paragraph{Constant and Time-Varying Noise Models.}\label{sec:noise}
Likelihood $p(\mobs | \muY, \varY)$ requires defining a noise process $\varY = \noiseprocess(\mx, \vz)$.  For the white-box model, we assume {\em constant variance} for each signal in $\mobs$; these are encoded using log-normal population-level variables.  For the black-box model, we take a non-parametric approach and learn a time-dependent variance model, using an additional black-box ODE model:
\begin{align}
\dot{\noisestate} &= \neuralfunction_3^{+}( \noisestate, \vx, \nnargs ) - \noisestate \odot \neuralfunction_4^{+}( \noisestate, \vx, \nnargs ) \label{eq:nnnoise}
\end{align}
In experiments, we solve $\noisestate$ simultaneously with $\vx$, which adds little computational overhead, but provides a flexible, data-driven approach to modelling observation noise.
This time-dependent variance process is applied element-wise as $\varY$ in Eq.~\ref{eq:likelihood}.  

\subsection{Simulation: Modified Euler}
For ODEs, we can explicitly write a solver to simulate the dynamical system.
Following \cite{chen2018neural}, if the ODE has too many parameters or the discretisation appears to affect the quality of the solution, we can learn gradients in a memory-efficient way through the adjoint method \cite{stapor2018optimization}.
The same simulation applies to white-box and black-box methods, and the difference is contained in the function $f_{\vtheta}$.

\section{Prior Work}
Prior work on learning the parameters of dynamical systems is extensive; we present an overview of key works here.
See Appendix~\ref{sec:extended_prior_work} for an extended discussion.

{\bf Markov-Chain Monte Carlo (MCMC)} methods have long been the gold standard for inference in ODEs (e.g., \citealt{xun2013parameter}).
\citet{dalchau2019} applied MCMC inference to the synthetic biological problem described in Section~\ref{sec:casestudy}, however reports convergence times of 24-48 hours. 
Due to numerical integration at each time step, MCMC is not scalable across large time series models, and lacks the flexibility, interpretability, and compositionality of our method.

{\bf Likelihood-free methods} (a.k.a. Approximate Bayesian Computation (ABC)) can infer parameters of dynamical systems without running a Markov chain to convergence.  For instance,  
Sequential Monte Carlo ABC simulates a discretised dynamical system to infer parameters of an ODE \citet{toni2009approximate}.
We choose instead to take advantage of fast, gradient-based optimization that scales better with data.

{\bf Gradient matching} avoids numerical integration by using a Gaussian Process (GP) regression to the state variables of the dynamical system.
\citet{gorbach2017scalable} use mean-field variational inference and gradient matching to learn parameters GPs that approximate dynamical systems.
GPs do not scale well to large datasets, and so are not applicable to our massive data regime.

{\bf Variational inference for dynamical systems} has been applied heavily in state-space models, e.g. \citet{archer2015black}.
Unlike \citep{archer2015black}, our method deals with dynamical systems that have hierarchical latent structure and highly {\em nonlinear} latent transitions.
Moreover, our model does not require full state observability, which is infeasible in many applications, including our case study.

\citet{krishnan2015deep} learn {\em nonlinear Kalman filters} through stochastic variational inference.
They explicitly generalise linear dynamical systems, discovering arbitrarily complex transition dynamics and emission distributions.
A key difference is that the mean and covariance functions for their (tridiagonal) variational distribution are recurrent neural networks.

\citet{ryder2018black} uses variational inference for {\em stochastic differential equations}.
Their approximate posterior factorizes into one component that determines the parameters for the SDE drift and diffusion matrices, and one that predicts the parameters of the latent process.
The component that describes the latent process evolution is autoregressive, requiring more computation.
Our method induces hierarchical structure in the approximate posterior rather than a complex noise process, and deals with dynamical systems in greater generality. 
Thus, given the generative process of an SDE, our method could also be used for Bayesian inference over such parameters.

\citet{chen2018neural} propose  {\em neural ordinary differential equations} and reinterpret deep neural networks that exhibit repeated composition as discretised ODEs.
They use a different modelling regime to ours, in that
we learn a block-conditional variational distribution over the parameters of a system of ODEs, while their
variational distribution is over the initial state of the latent time series, with the goal of reinterpreting deep learning models as continuous-time dynamical systems.
We work from the opposite direction, inducing hierarchical structure in the variational distribution for more efficient sharing of statistical strength during parameter identification.

\section{Synthetic Biology Case Study}
\label{sec:casestudy}

\newcommand{\rfp}{RFP}
\newcommand{\cfp}{CFP}
\newcommand{\yfp}{YFP}

We used our methodology to model bacterial cells that were genetically engineered to respond to specific input signals. Cells are typically engineered by inserting DNA that instructs the cells to produce proteins, which can implement a range of information processing behaviours, including  Boolean logic gates \cite{nielsen2016}, analog computation \cite{daniel2013}, or inter-cellular communication \cite{grant2016}. 

\begin{figure}[ht]
	\includegraphics[width=\linewidth]{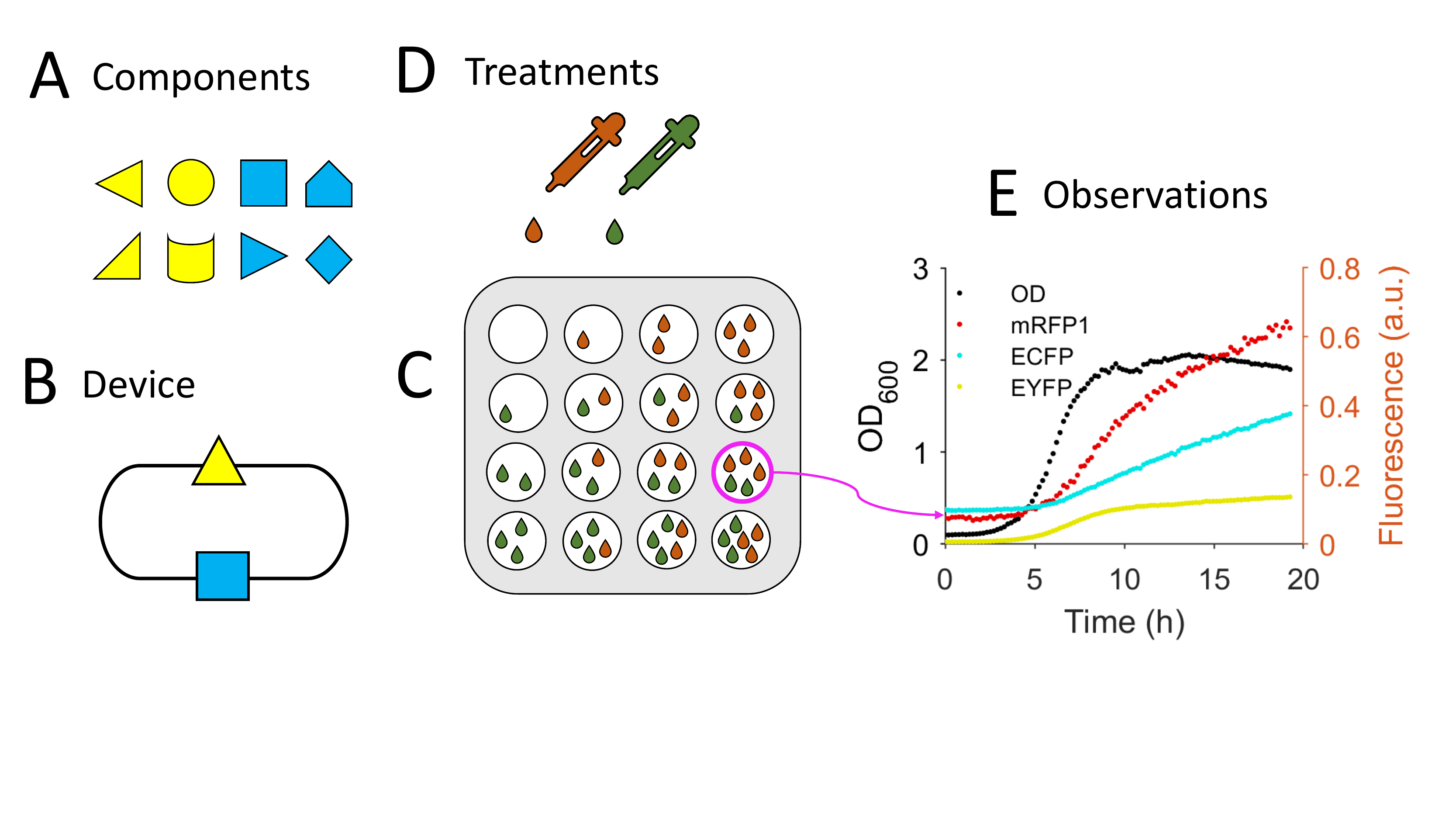}
	\vspace{-0.6in}
	\caption{ {\bf Schematic view of the Synthetic Biology Case Study}    \textbf{A}: Exchangeable genetic components are selected and inserted into a circular piece of DNA called a \textit{plasmid} (\textbf{B}), which genetically encodes a \textit{receiver device}.  Each receiver device contains a different combination of two types of components (yellow and blue), each with several variants. \textbf{C}: The plasmid is inserted into a bacterial cell culture, which is replicated and grown in wells (circles) on an experimental plate.  \textbf{D}: Each well is treated with different concentrations of chemicals $C_6$ and $C_{12}$ and grown for approximately 24 hours.  \textbf{E}: During the growing period cameras capture OD, RFP, YFP, CFP.}
	\label{fig:casestudy}
\end{figure}

We considered a population of bacterial cell cultures, each with a different combination of shared genetic components (Figure~\ref{fig:casestudy}). 
Each cell culture contained genetic components implementing a \textit{double receiver} device, which functions as a biosensor by producing two \textit{receiver proteins}, $R$ and $S$. 
Receiver $R$ binds to input signal $C_6$ to activate production of cyan fluorescent protein (CFP), and receiver $S$ binds to input signal $C_{12}$ to activate production of yellow fluorescent protein (YFP) \cite{grant2016}. 
Genetic components associated with the efficiency of $R$ production were one of (Pcat, RS100, R33) and equivalently one of (Pcat, S32, S175, S34) for $S$.  
Each culture continuously produces a red fluorescent protein ($\it{RFP}$), providing a baseline for cellular activity.
Finally, the size of the culture is measured as optical density (OD). 
\vspace{-0.1in}
\paragraph{White-box model}
The white-box ODE model we consider describes the evolution of the density $c$ of the cell culture as $\dot{c} = \gamma(c).c$.
The function $\gamma(c)$ defines a lag-logistic scheme, considering the growth rate to be 0 before a time $t_\text{lag}$ is reached, at which point there is a transition to logistic (density-limited exponential) growth. 
\begin{equation}
\gamma(c) = \textcolor{\indcolor}{r}.\left(1-\nicefrac{c}{\textcolor{\indcolor}{K}}\right)/(1 + e^{-4(t-\textcolor{\indcolor}{t_{\text{lag}}})}) 
\label{eq:gamma_smooth}
\end{equation}
Additional state variables describe the intracellular concentrations (denoted $[\cdot]$) of receiver device components RFP, YFP, CFP, LuxR ($R$) and LasR ($S$), and two additional variables quantifying cellular autofluorescence at 480~nm ($F_{480}$) and 530~nm ($F_{530}$).
\begin{align*}
\dot{[\text\rfp]} &= r_c - (\textcolor{\popcolor}{d_\text\rfp} + \gamma).[\text{\rfp}] \\
\dot{[\text\cfp]} &= \textcolor{\popcolor}{a_\text\cfp}.r_c.f_{76}(C_6,C_{12},[R],[S]) - (\textcolor{\popcolor}{d_\text\cfp}+\gamma).[\text\cfp] \\
\dot{[\text\yfp]} &= \textcolor{\popcolor}{a_\text\yfp}.r_c.f_{81}(C_6,C_{12},[R],[S]) - (\textcolor{\popcolor}{d_\text\yfp}+\gamma).[\text\yfp] \\
\dot{[R]} &= \textcolor{\groupcolor}{a_R}.r_c - (\textcolor{\popcolor}{d_R} + \gamma).[R] \\
\dot{[S]} &= \textcolor{\groupcolor}{a_S}.r_c - (\textcolor{\popcolor}{d_S} + \gamma).[S] \\
\dot{[F_{480}]} &= \textcolor{\popcolor}{a_{480}}.r_c - \gamma.[F_{480}] \\
\dot{[F_{530}]} &= \textcolor{\popcolor}{a_{530}}.r_c - \gamma.[F_{530}]
\end{align*}
Here, $f_{76}$ and $f_{81}$ are the derived functional responses to input signals and receiver proteins, as defined in \citet{grant2016} (also see Appendix~\ref{sec:mechanistic}). 
The parameter $r_c$ represents the metabolic activity of the culture, which can vary between cultures, and is therefore an individual parameter.
We assume that the growth parameters are also specific to a given culture, and so $\vz_\indvar = \{r,K,t_\text{lag},r_c\}$.
The receiver devices in each culture differed by two genetic components, $a_R$ and $a_S$, responsible for the production of the two receiver proteins $R$ and $S$, respectively, such that $\textcolor{\groupcolor}{\vz_{\groupvar}=\{a_R,a_S\}}$. 
All remaining parameters {\color{blue}$\vz_{\popvar}$} are shared across all cell cultures.
The experiment conditions $\vtreat = [C_6,C_{12}]$ are assumed to be constant throughout the course of an experiment. 

\paragraph{Observer process}
The observer process $\observerprocess$ defines how to relate the model to the ($\nspecies=4$) measured signals.
Optical density (OD) represents the density of the culture, and is modelled explicitly by the white-box model.
By analogy, we impose that the first internal state of the black-box model ($x_0$) models cell density.
\emph{Bulk} fluorescence measurements of the whole culture are then collected at excitation wavelengths corresponding to RFP, YFP and CFP.
In the white-box model, contributions to bulk fluorescence includes fluorescent proteins and autofluorescence as separate variables; scaling the sum of the protein and autofluorescence concentrations by the cell density provides an approximation of bulk fluorescence for each signal.
However, in the black-box model, internal states cannot be separated easily under addition, so we use single variables each to represent intracellular RFP, YFP and CFP, and do not explicitly separate autofluorescence.
Accordingly, the observer processes for each model differ in structure slightly (Table \ref{tab:observer}).
\begin{table}[ht] \centering
	\caption{The observer process for white-box and black-box models.}
	\begin{tabular}{ccc}
		\toprule
		Signal & White-box & Black-box \\
		\midrule
		OD & $c$ & $x_0$ \\
		RFP & $c.[\text\rfp]$ & $x_0 . x_1$ \\
		YFP & $c.([\text\yfp]+[F_{530}])$ & $x_0 . x_2$ \\
		CFP & $c.([\text\cfp]+[F_{480}])$ & $x_0 . x_3$ \\
		\bottomrule
	\end{tabular}
	\label{tab:observer}
\end{table}
  
\section{Experiments}\label{sec:experiments}
We performed two experiments on data from the synthetic biology case study, where measurements of six genetic devices (Pcat-Pcat, RS100-S32, RS100-S34, R33-S32, R33-S175 and R33-S34) were combined into a collection of 312 time-series.  
In the first experiment, we split the dataset into cross-validation folds and compared a block-conditional white-box model (see Section~\ref{sec:casestudy}) with a black-box model using biologically plausible constraints on its functional form (see Section~\ref{sec:decoder}).  
In the second experiment, the combined dataset had one device held out as a test dataset, while the remaining devices underwent cross-validation for model selection.  
This experiment was performed on devices whose components had not appeared in combination together in the training set, but had appeared separately in other devices, enabling their parameters to be identified.
The ability of the method to predict devices not previously seen before is particularly compelling when seeking to select components that, when put together, implement a given specification.

\subsection{Architecture and Optimisation Details}   
For $q(\vz_{\indvar} | \mobs, \vgroup)$, we use the same encoder NN for both white-box and black-box models: 10 1D convolutional filters, feeding 50 unit hidden layer with tanh activations; $\vgroup$ is concatenated to the hiddens, which is then connected to the mean and variances outputs of $q$. Neural networks $\neuralfunction^{+}$ all have $25$ hidden units.  All decoder NNs are the same for all black-box models.  During training we used a K=100 importance weighted auto-encoder (IWAE) estimator for gradient computation \citep{burda2015importance}.  
We implemented a doubly reparameterised gradient estimator (DReG) for IWAE, which has recently been shown to yield lower-variance gradient estimates than the gradient computed naively by reverse-mode automatic differentiation \citep{tucker2018doubly}, while also avoiding the known problem of poor gradients for the variational approximation with greater numbers of importance samples \citep{rainforth2018tighter}.  We found that the DReG estimator gave modest improvements for the black-box, but {\em very significant improvements} in performance and rate of convergence for the white-box (see Appendix figure~\ref{fig:elbo}).  For evaluation, we used $K=1000$ IWAE samples.  We implemented a modified-Euler solution to simulate our ODEs; this was implemented in tensorflow using tf.while blocks.  We ran all experiments for $500$ epochs using Adam optimisation \cite{kingma2014adam}.  To perform 4-fold cross-validation, black-box models take approximately $40$ minutes and white-box models approximately $2$~hrs. This is significantly faster than MCMC (see Section~\ref{sec:mcmc}), which required approximately 40 times more computation.    

\subsection{Multiple Device Inference}

\begin{figure}[ht]
	{\sf A} \\
	\includegraphics[width=\linewidth]{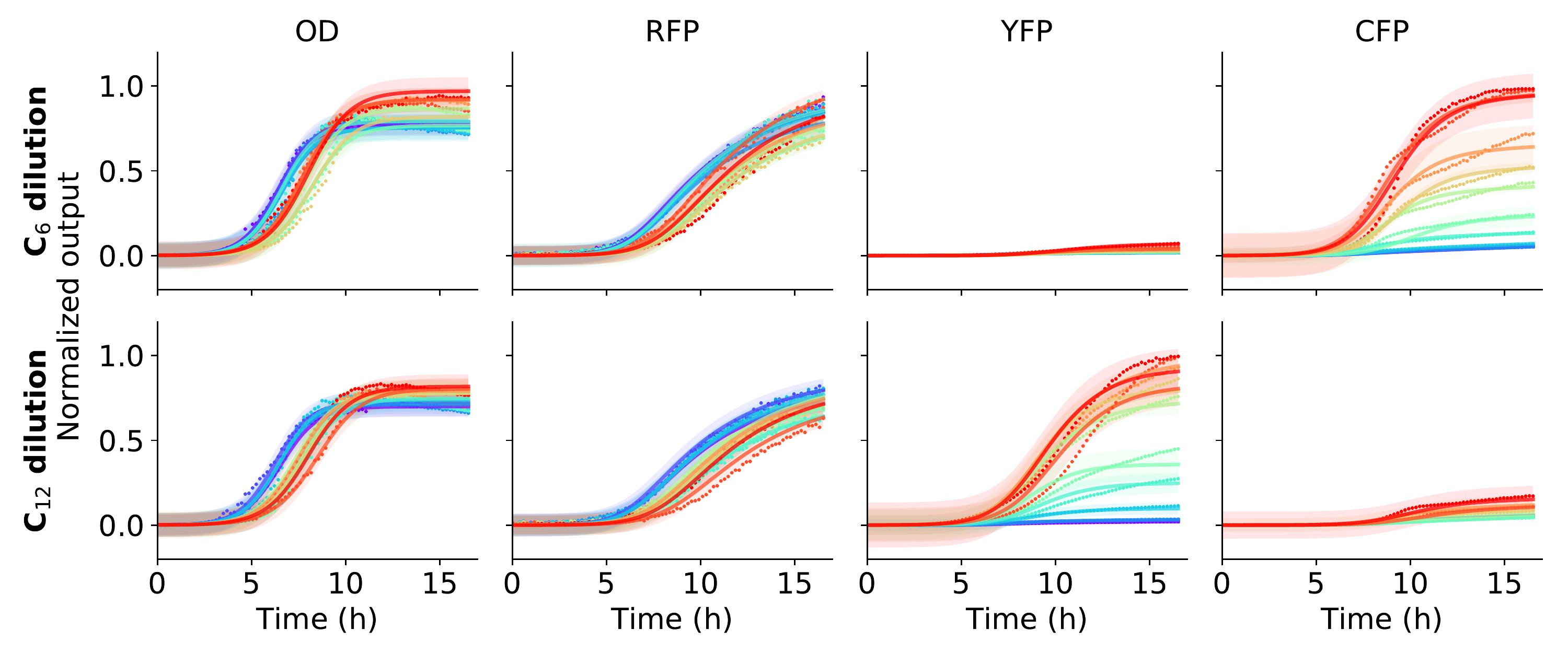}
	{\sf B} \\
	\includegraphics[width=\linewidth]{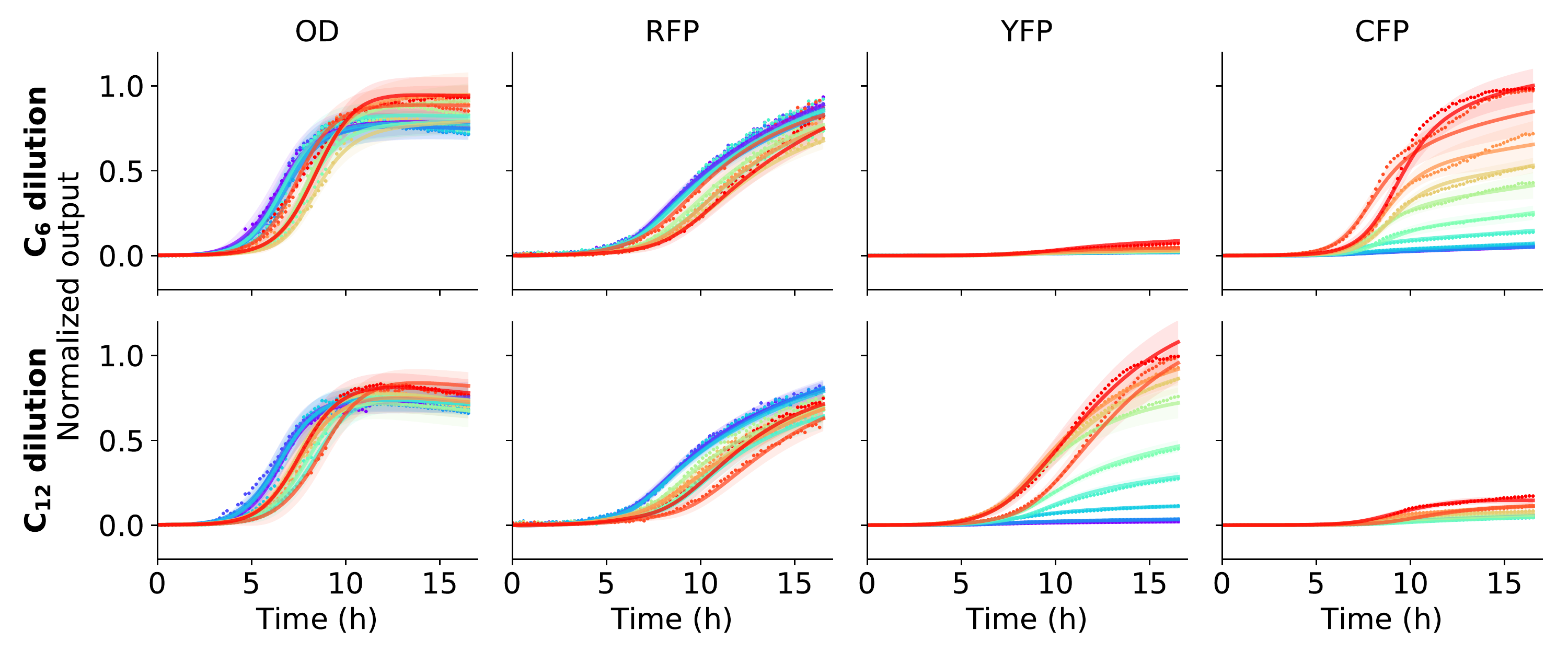}
	\vspace{-0.3in}
	\caption{\textbf{Posterior predictive distribution for the R33-S175 device.} (\textbf{A}) White-box model. (\textbf{B}) Black-box model. For clarity, we selected one of each replicated treatment, so that a full dilution series can be visualised for $C_6$ and $C_{12}$, sweeping from low to high YFP/CFP expression. Predictions for the full dataset are shown in the Appendix (Figure~\ref{fig:xval}).}
	\label{fig:postpredict1}
\end{figure}

\begin{figure*}[ht]
	\includegraphics[width=\linewidth]{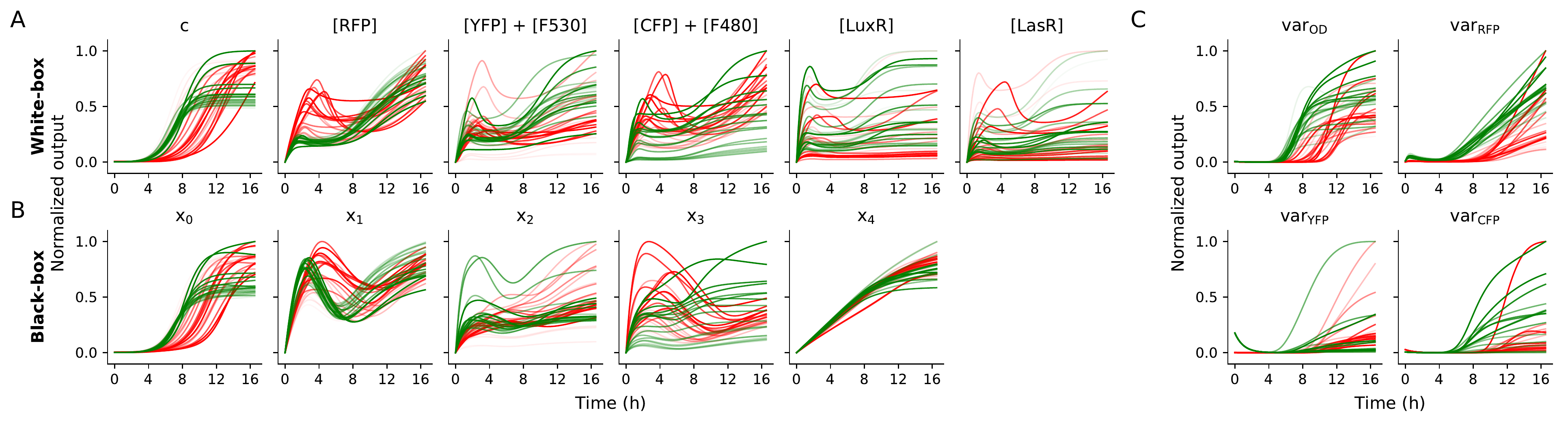} 
	\vspace{-0.3in}
	\caption{\textbf{Predicted dynamics of the state variables.} The internal state variables for simulation of R33-S34 in the cross-validation experiment with the (\textbf{A}) white-box and (\textbf{B}) black-box models are compared. Simulations are separated by whether they correspond to $C_6$ treatments (green) or $C_{12}$ treatments (red), with lower concentrations indicated by lighter shades. (\textbf{C}) Dynamics of internal states used for the time-varying noise of the black-box model.}\label{fig:latents}
\end{figure*}

We first ran 4-fold cross-validation with 500 epochs and batch size 36, and then collected the posterior predictive distributions for the cross-validated portions of each fold.  
Both white-box and black-box models were able to capture the majority of the dynamical behaviour (Figures~\ref{fig:postpredict1}, \ref{fig:xval}--\ref{fig:individual_R33S175}).
In Table~\ref{tab:xvalelbos} cross-validation  ELBO estimates are presented for variety of latent variable and ODE signal capacity for the black-box.  Since the encoder neural networks are fixed for both models, we only report the single estimated ELBO for the white-box.  Black-box has two significant advantages that explain the improved performance: 1) it has flexible modelling capacity in the form of its neural differential equations, and 2) it has a more suitable and realistic neural noise process, enabling the posterior predictive distribution to exhibit lower variance, especially at earlier time points. In Figure~\ref{fig:latents}C we plot the dynamic noise process for the black-box model. 

\begin{table}[ht]
    \caption{ {\bf Black-box Cross-validation Results}.  For comparison, the white-box cross-validation ELBO is 1047 nats. $M_x$ is the number of species in the black-box, where the first 4 are used in the observer process, so $M_x=5$ means there is $1$ extra {\em latent species}. For the figures we select the black-box model $\star$, but we note no significant difference between results shown italics.  } 
    \centering
    \begin{tabular}{ccc|cccc}
    \toprule
    \multicolumn{3}{c|}{\# Variables} & \multicolumn{4}{|c}{Mean IWAE ELBO, K=1000} \\ %\hline 
        \popvar & $\groupvar$ & $\indvar$ & $M_x=4$ & $5$ & $6$ & $8$  \\ \midrule
        2 & 0 & 0 & 1155 & 1215 & 1210 & 1213  \\  %\hline
        2 & 2 & 0 & 1201 & 1277 & 1280 & 1283  \\  %\hline
        2 & 2 & 2 & 1175 & {\em 1352} & {\em 1340} & {\em 1340}  \\  %\hline
        5 & 2 & 5 & 1204 & {\bf 1360}$^\star$ & 1321 & 1305 \\ %\hline   
        \bottomrule
    \end{tabular}
    \label{tab:xvalelbos}
\end{table}

\begin{figure}[ht]
	{\sf A ~~~ {\scriptsize Multiple Device Inference}} \\
	\includegraphics[width=\linewidth]{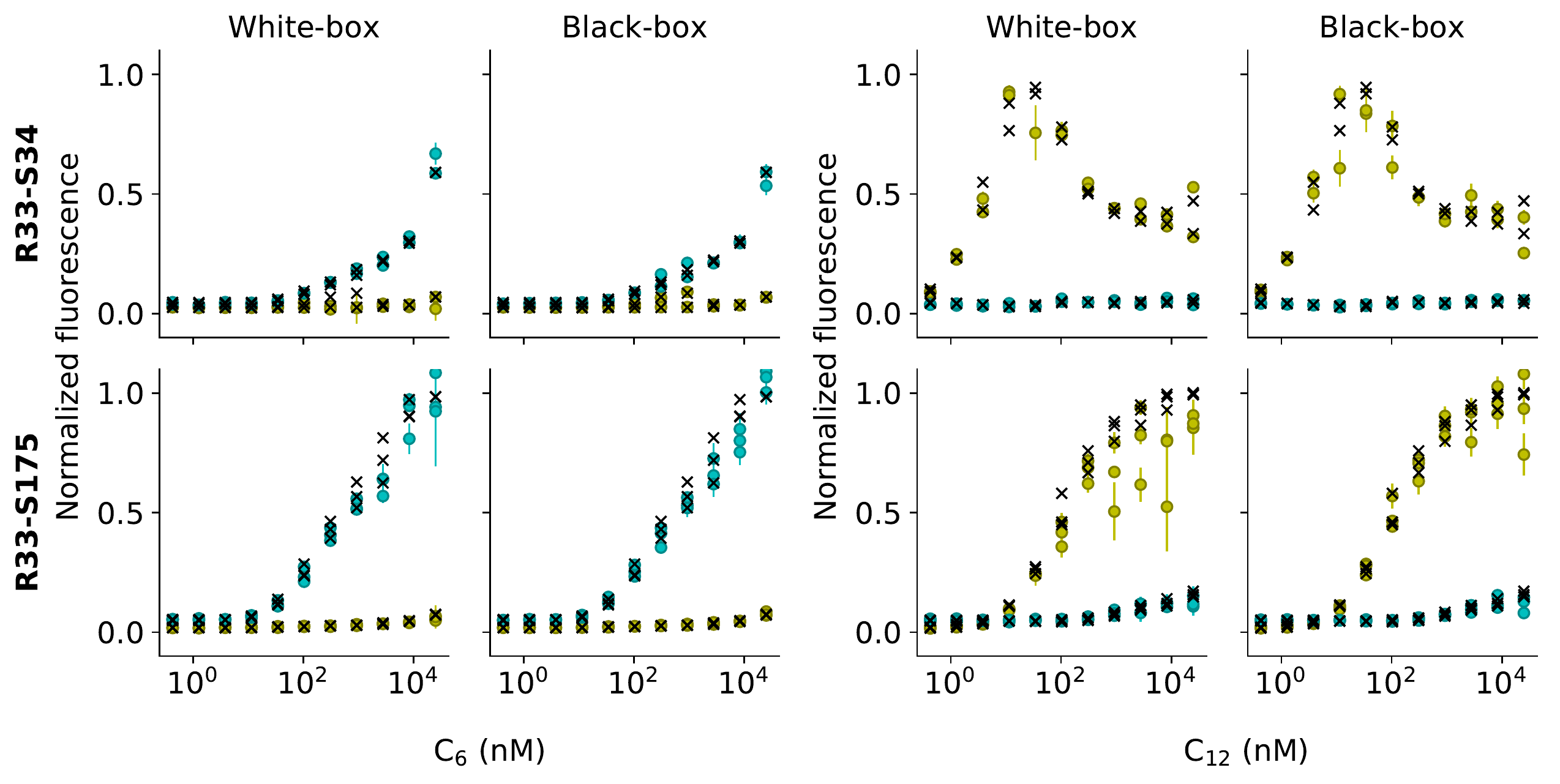} \\
	{\sf B ~~~ {\scriptsize Held-Out Device Inference}} \\
	\includegraphics[width=\linewidth]{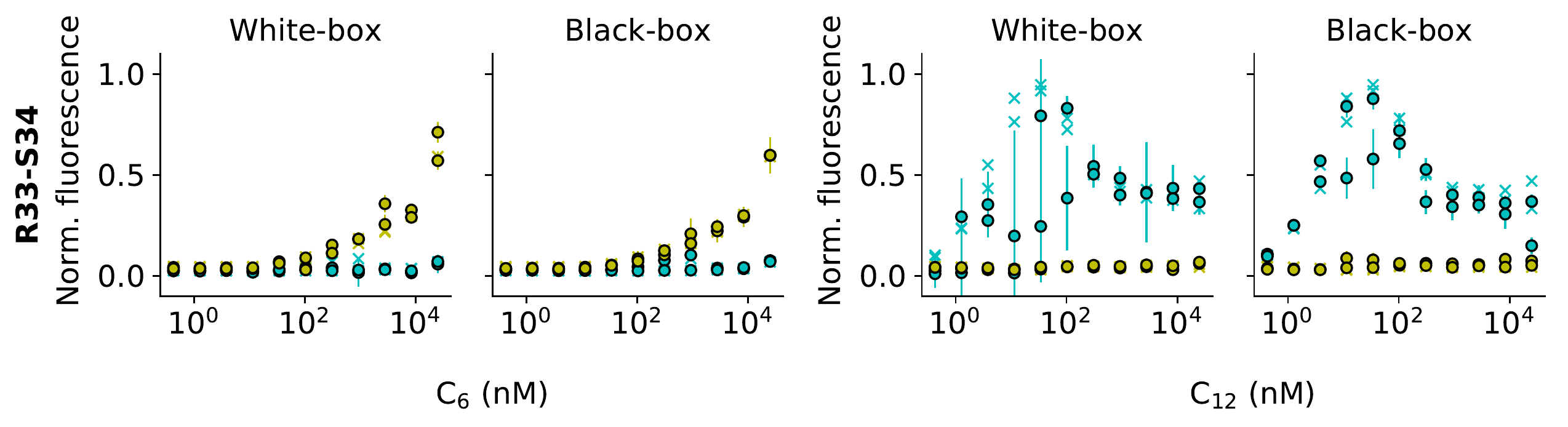} 
	\vspace{-0.3in}
	\caption{\textbf{Predicted model response to input signals.} (\textbf{A}) 4-fold cross-validation inference and (\textbf{B}) Held-out device inference. To summarise how each model predicts the overall response of R33-S34 and R33-S175 to input signals, we show the posterior predictive distribution at the final time-point (circle with error bars denoting two standard deviations), compared against the data point (crosses) for YFP (yellow) and CFP (cyan).}
	\label{fig:postpredict2}
\end{figure}

To provide a summary view of how each model predicts the response to the input signals $C_6$ and $C_{12}$, we plotted the posterior predictive distributions at the final time-points as a function of the input concentrations.
For the R33-S175 device, the responses of CFP to  $C_6$ and YFP to $C_{12}$ are both monotonically increasing.
These responses are well-predicted by both models (Figures~\ref{fig:postpredict2}A, \ref{fig:treatments}), though the black-box model captured the actual measured values with higher precision.
In contrast, the R33-S34 device exhibits a non-monotonic shape of YFP expression in response to $C_{12}$. 
Because the transfer function hypothesised to describe the equilibrium response of YFP production ($f_{81}$, see Appendix) is monotonic increasing with respect to $C_{12}$, it would seem that it cannot explain the observed trends for R33-S34 in bulk fluorescence. 
Here we find that this can be explained by high $C_{12}$ concentrations slowing down cell growth (Figure~\ref{fig:individual_R33S34}).
While the intracellular concentration of YFP is increased by the monotonic $f_{81}$, lower cell densities produce lower bulk fluorescence.
While there is low precision in the predictions at the final time points of high $C_{12}$ treatments for the white-box model, it captures the general trend almost as well as the black-box model.
This valuable modelling insight was really made possible by the simultaneous inference of multiple devices and multiple signals. \\

\vspace{-0.1in}

Using the black-box approach, we can overcome model misspecification, producing accurate predictions across all 6 devices. 
However, the interpretability of these models is more restricted.
Nevertheless, we made an effort to enforce plausibility constraints on the state variables of the black-box model, both through separating production and degradation terms (ensuring non-negativity of the state variables), and using a similar observer process.
As such, comparison of the dynamics of the state variables representing intracellular concentrations in the black-box model might inform what is inaccurate in the white-box model. 
For example, during the first 4 hours following treatment by $C_6$/$C_{12}$, the white-box model has more accentuated transient variations in [\yfp] and [\cfp] (Figures~\ref{fig:latents}A, \ref{fig:latentsWB}) than equivalent state variables in the black-box model (Figures~\ref{fig:latents}, \ref{fig:latentsBB}).

\subsection{Held-out Device Inference}
\label{sec:held_out}

We explored the more challenging scenario of predicting data for devices held-out from the training set. 
When holding out the R33-S34 device, both the white-box and black-box models were able to accurately capture the moderate increasing response of CFP and almost absent response of YFP to increasing concentrations of $C_6$ (Figure~\ref{fig:postpredict2}B). 
The predictions of the YFP response to $C_{12}$ by the white-box model were imprecise, indicating that either more training data is required to make clearer predictions, that over-fitting has occurred, or that inherent model misspecification is limiting the ability to predict accurately.
But as the black-box model was able to correctly predict the observed non-monotonic response with moderate precision, we can rule out the lack of training data as a cause.
In general, the ability to compare black-box and white-box models in this way is useful for providing an upper bound on what might be achievable by any prescribed model.

\section{Discussion and Future Work}\label{sec:discussion}
We have introduced {\bf \name}, a framework for learning and inference of hierarchical and nonlinear dynamical models, and applied it to a careful set of experiments on a synthetic biology case study. \name~ brings several elements from recent machine learning advances together into one framework with the following key advantages: efficient amortised inference using VAEs, multilevel latent variable modelling with block-conditional factorisation; a broad range of dynamic behaviour modelling, from  user prescribed white-box dynamics to black-box dynamics with neural sub-components.  

Although the experiments for our synthetic biology case study focused on ODEs--already a broad class of dynamical systems--our framework could be further generalised to include stochasticity in the dynamics model, for both the white-box and black-box cases. We could potentially avoid sampling altogether by adopting deterministic VI \citep{wu2018deterministic} to propagate the variational posterior all the way through the ODE solver to the likelihood.  Treating inference as optimisation opens the door to novel active learning experiments for dynamical systems of synthetic biology, where additional cost functions provide gradients to maximise desirable properties, similar to work in synthetic chemistry \citep{gomez2018automatic}.   

\section*{Acknowledgments}
Geoffrey Roeder is partially supported by the Natural Sciences and Engineering Research Council of Canada (NSERC), funding reference number PGSD3-518716-2018.
\bibliography{refs}
\bibliographystyle{icml2018}

 \clearpage
\appendix
\section*{Appendix}
\section{Extended Case Study Description}
\label{sec:extended_prior_work}

The synthetic genetic circuits we have used in this work are built from gene cassettes comprised of DNA sequences encoding a promoter, ribosome binding site, coding region, and terminator. 
These cassettes are assembled into plasmids that are used to transform \textit{E. coli} cells. 
We refer to a collection of cassettes that implement a particular design as a \emph{device}.  
The devices used in this paper are all \textit{double receiver} devices that respond to 3-oxo-C6-homoserine lactone (C6) by producing cyan fluorescent protein (CFP) and to 3-oxo-C12-homoserine lactone (C12) by producing yellow fluorescent protein (YFP) \cite{grant2016}.  
These devices are built of 5 cassettes.
The first 4 cassettes are arranged on a plasmid, which includes one cassette each for producing luxR and lasR proteins ($R$ and $S$ in the main text), the receiver proteins that bind C6 and C12, respectively, a CFP cassette activated by C6-bound luxR, and a YFP cassette activated by C12-bound lasR.
The fifth cassette is chromosomally integrated, and constitutively expresses RFP.
The devices vary in the strength of the ribosome binding sites in the luxR and lasR cassettes, creating devices that vary in the amount of each of those proteins expressed and therefore their sensitivity to C6 and C12. 

\subsection{White-box (mechanistic)}
\label{sec:mechanistic}

Our general approach for constructing prescribed (white-box) models of biological circuits combines a population-level model for cell culture growth with more detailed models for the concentrations of intra- and intercellular molecules, and resembles the approach commonly used in the synthetic biology literature \cite{balagadde2008,daniel2013,chen2015,dalchau2019}.
Cell growth models are generally described by the product of the current cell density $c(t)$ and the \emph{specific growth rate} $\gamma(c(t))$, which describes both the per capita growth rate and the decrease in intracellular concentrations due to an increased volume.
As explained in the main text, we used a smoothed version of the lag-logistic model for cell growth here.

To model the cellular biochemistry, we translate chemical reaction networks to ODEs using mass action kinetics, which assumes that reactions fire at a rate proportional to the concentration of the reactants.
Translating chemical reactions in this way in general leads to a large number of equations, because all mRNAs, proteins, small molecules and complexes between each produce their own equation.
As such, model reductions are commonly applied to reduce the number of dependent variables, but result in more complex nonlinearities. 
Following this approach, the white-box model we consider here (Section~\ref{sec:casestudy}) was derived in detail previously \cite{grant2016,dalchau2019}. 
It describes the time-evolution of the response of double receiver devices to HSL signals $C_6$ and $C_{12}$, vector $\vu$ in \eqref{eq:ode}. 
The latent variables $\vx$ in \eqref{eq:ode} are the culture density $c$, the intracellular concentrations of each expressed protein (luxR, lasR, RFP, CFP, YFP) and variables for autofluorescence, which we model as concentration of intracellular material fluorescent at 480 nm (F$_{480}$) and 530 nm (F$_{530}$).

As there are no mRNA species, the variables $a_k$ describe a lumped maximal rate of transcription and translation. 
The variables $d_k$ describe the intracellular degradation rates of each protein. 

The \emph{response} functions $f_{76}$ and $f_{81}$ describe the inducibility of CFP and YFP to complexes involving the HSL signals and the receiver proteins luxR and lasR. 
The response functions were derived from chemical reactions, making the assumption that signal-receiver binding and unbinding is faster than protein synthesis and degradation \cite{dalchau2019}.
This results in very complex functions, but they are still interpretable as exhibiting saturation behaviour, which occurs as either receiver proteins or promoters become limiting.
We define $B^{(k)}_R$ and $B^{(k)}_S$ as the fractions of luxR and lasR proteins bound by an HSL signal
\begin{subequations}
	\begin{align}
	B_R &= \dfrac{(K_{R6}.C_6)^{n_R} + (K_{R12}.C_{12})^{n_R}} {(1 + K_{R6}.C_6 + K_{R12}.C_{12})^{n_R}} \\
	B_S &= \dfrac{(K_{S6}.C_6)^{n_S} + (K_{S12}.C_{12})^{n_S}} {(1 + K_{S6}.C_6 + K_{S12}.C_{12})^{n_S}},
	\end{align}
\end{subequations}
These functions are bounded above by 1, which occurs when luxR/lasR become limiting.
The CFP or YFP genes are transcribed more efficiently when their promoters are bound by one of the receiver-signal complexes. 
As such, an additional saturation can be observed within the derived forms
\begin{equation} \small
f_j(R,S,C_6,C_{12}) = \dfrac{\epsilon^{(j)} + K_{GR}^{(j)} R^2 B_R + K_{GS}^{(j)} S^2 B_S} 
{1 + K_{GR}^{(j)} R^2 B_R + K_{GS}^{(i)} S^2 B_S}
\end{equation}
where $j\in\{76,81\}$.
Here, the parameters $K_{GR}^{(j)}$ and $K_{GS}^{(j)}$ are the affinity constants of receiver-signal complexes for each promoter $j$, and $\epsilon^{(j)}$ is the leaky rate of transcription/translation in the absence of an activating complex (such as when there is no HSL).

The specific growth rate $\gamma(c_i)$ describes the per-capita cellular growth rate of culture $i$. 
Cultures are subscripted in this way because their growth parameters will be local to the culture, which enables implicit accounting for feedback from circuit activity or extrinsic factors that vary in different experiments.
As in \cite{dalchau2019}, we use a lag-logistic growth model which explicitly quantifies a \emph{lag} phase of bacterial growth before an  \emph{exponential} phase and then \emph{stationary} phase.
This is usually formulated as
\begin{equation*}
\gamma(c_i) = \begin{cases} r_i.(1-\frac{c}{K_i}), & t>t_{\text{lag},i} \\ 0, & t<t_{\text{lag},i}\end{cases}
\end{equation*}
but to ensure that the right-hand sides of $f$ are differentiable, we replace the switch at $t_{\text{lag},i}$ with a steep sigmoid (Equation~\ref{eq:gamma_smooth}).

Finally, we remind the reader that we consider the application of this model to multiple devices, in which different ribosome-binding site sequences have been used for the luxR and lasR genes (Section \ref{sec:casestudy}).
In this mechanistic model, it is the parameters $a_R$ and $a_S$ that are allowed to be device-conditioned. 
Therefore, using the one-hot mapping strategy for specifying the rbs elements in each device, the model can take on device-specific quantities for luxR and lasR synthesis.

\section{Encoder Architecture}

\paragraph{Local random effects.}

The parameters ${\vphi}_{local}$ consist of real scalar functions $m$ and $s$ such that 
\begin{align*}
q_{\vphi_{local}}^{(i)} = \mathcal{N} \left( m(\vy, \vu, \vo; \xi_i), s(\vy, u, \vo; \eta_i) \right)
\end{align*}
The $m$ and $s$ functions have the following form.
In the following explication, take $\vy$ and $u$ to be arbitrary but particular samples from the true data-generating distribution.

We first apply a 1D strided CNN to the input $\vy$ s.t.
$\tilde{\mathbf{y}} = \texttt{flatten}( \texttt{AvgPool}( \texttt{ReLU}( \vy * W_1 ) ) )$, where the convolution filter $W_1$ is chosen so that $\vy * W_1$ has one output channel.
Note that the width of the convolution filter influences the correlation structure we can discover in the time series, acting as a window across the sequence.
This stage in the computational pipeline is designed for feature learning.

Next, $\tilde{\vy}, u, \text{and } \vo$ are concatenated as inputs to an MLP with $\tanh$ nonlinearity and no bias term, that is, $h = \tanh( W_2 \cdot [\tilde{\vy}; u; o] )$.
The composition of the two stages is a differentiable map $m := h \circ \tilde{\vy}$ s.t. $m: \mathbb{R}^{4T + n_u + n_d} \rightarrow \mathbb{R}$, yielding a mean for the learned posterior distribution over that parameter of the ODE.

The $s$ function is identical to the $m$ function but with different parameters, and it further exponentiates the output of the MLP.
This last mapping is equivalent to interpreting the unconstrained output of the second neural network as the log of the standard deviation.
We require the final step for $m$ to ensures that the resulting covariance matrix of the diagonal Gaussian is PSD.
Each layer of $m^{(i)}$ and $s^{(i)}$ is L2-regularized using independent constants $\lambda_j^{i}$ to counter overfitting during learning.

Over minibatch learning, the network parameters that map data and conditioning information to means and log standard deviations are adapted through gradient descent, achieving amortization and supporting generalization.

\paragraph{Global random effects.}
The parameters $\vphi_{global}$ are independent of the input data, and are real scalars $r$ and $v$ such that 
\begin{align*}
q_{\vphi_{global}}^{(i)} = \mathcal{N} \left( r_i, \exp \{v_i\} \right),
\end{align*}
that is, we optimize $v_i$ in an unconstrained space interpreting, it as the log standard deviation.
The variables $r_i$ and $v_i$ are initialized to the prior, and then optimized through gradient descent to maximize the ELBO. 

The prior was set in consultation with the domain experts who designed and conducted the experiments, collected the data, and defined the system of differential equations to capture reasonable upper and lower ranges of the parameters of interest.
We note that this could also have been achieved by empirical Bayes or Type-II Maximum Likelihood, which involves learning a hyperprior over the prior parameters, but have left this to future work given the availability of expert knowledge.

Each $q_{global}^{(i)}$ is a 1D Gaussian or LogNormal (e.g., its output is exponentiated) depending on whether the term it represents in the ODE is a strictly positive real-world quantity or it represents any real value.

% !TEX root = supplement.tex\\
\section{Comparison with MCMC}

To determine how VI-HDS compares with a simple approach to Bayesian inference, we sought to approximate the inference problem using Markov Chain Monte Carlo (MCMC). In principle, MCMC methods can provide an exact characterisation of the posterior, but often many samples are necessary for convergence. 
For comparison with our VI-HDS results, we generated MCMC chains for the white-box model. 
Due to the presence of 4 \emph{individual} parameters in the model ($r$, $K$, $t_\text{lag}$, $r_c$), and $N=312$ data-points, there were more than $4\times312$ parameters to be sampled. 

\label{sec:mcmc}
\begin{figure}[ht!]
	\includegraphics[width=\linewidth]{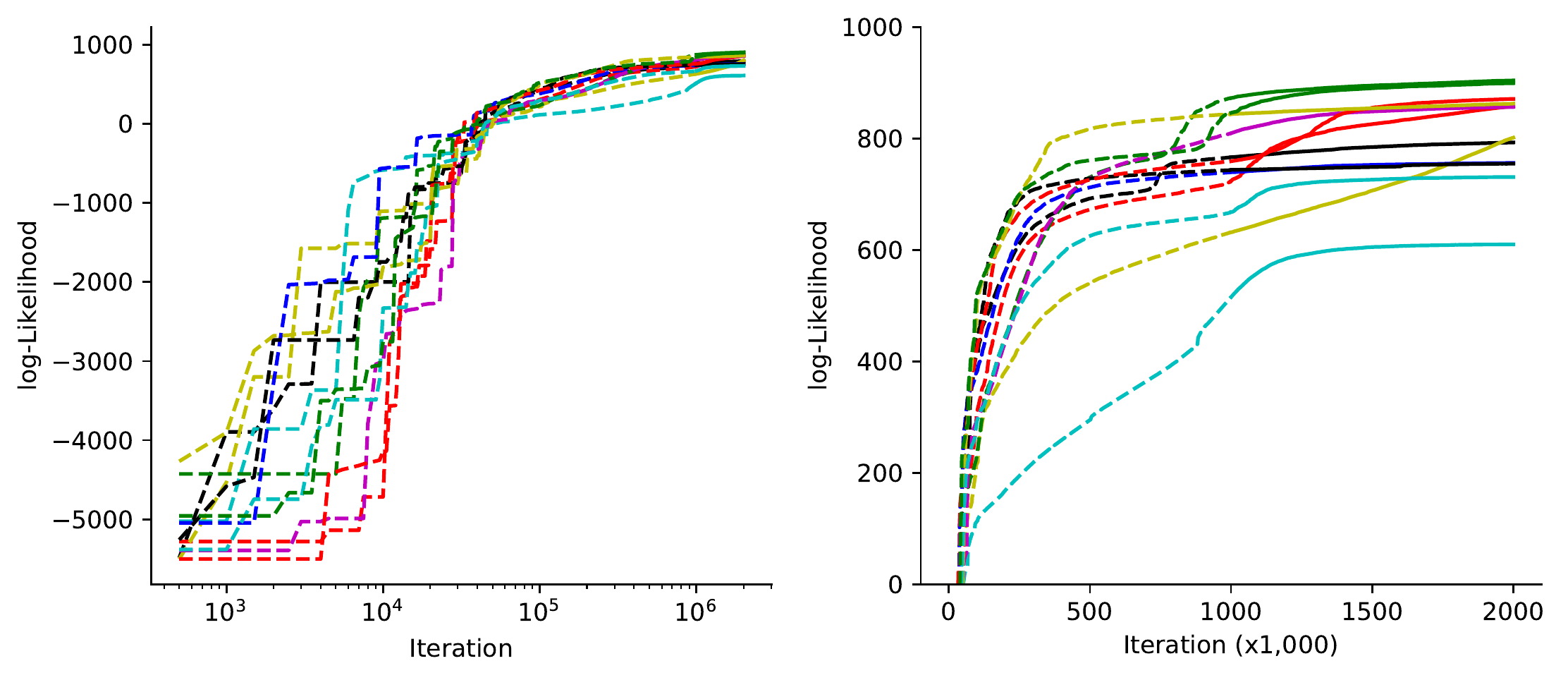}
	\caption{Convergence of Markov chain Monte Carlo}
	\label{fig:mcmc_chains}
\end{figure}
With so many parameters, we found that even 1 million burn-in samples was insufficient to reach convergence (Figure~\ref{fig:mcmc_chains}). 
Furthermore, local optima of the likelihood function were difficult to avoid.

\begin{figure}[ht!]
	{\sf A} \\
	\includegraphics[width=\linewidth]{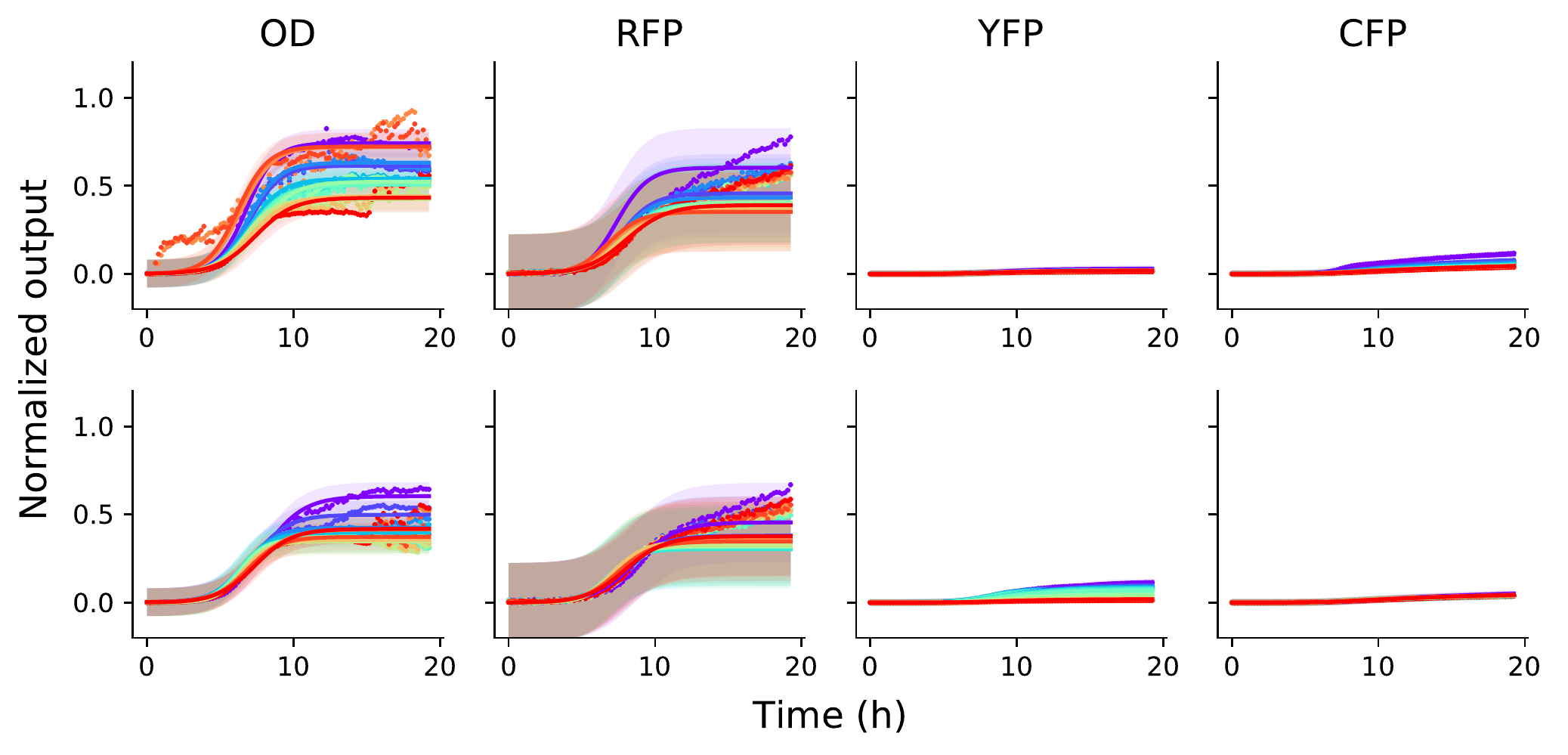} \\
	{\sf B} \\
	\includegraphics[width=\linewidth]{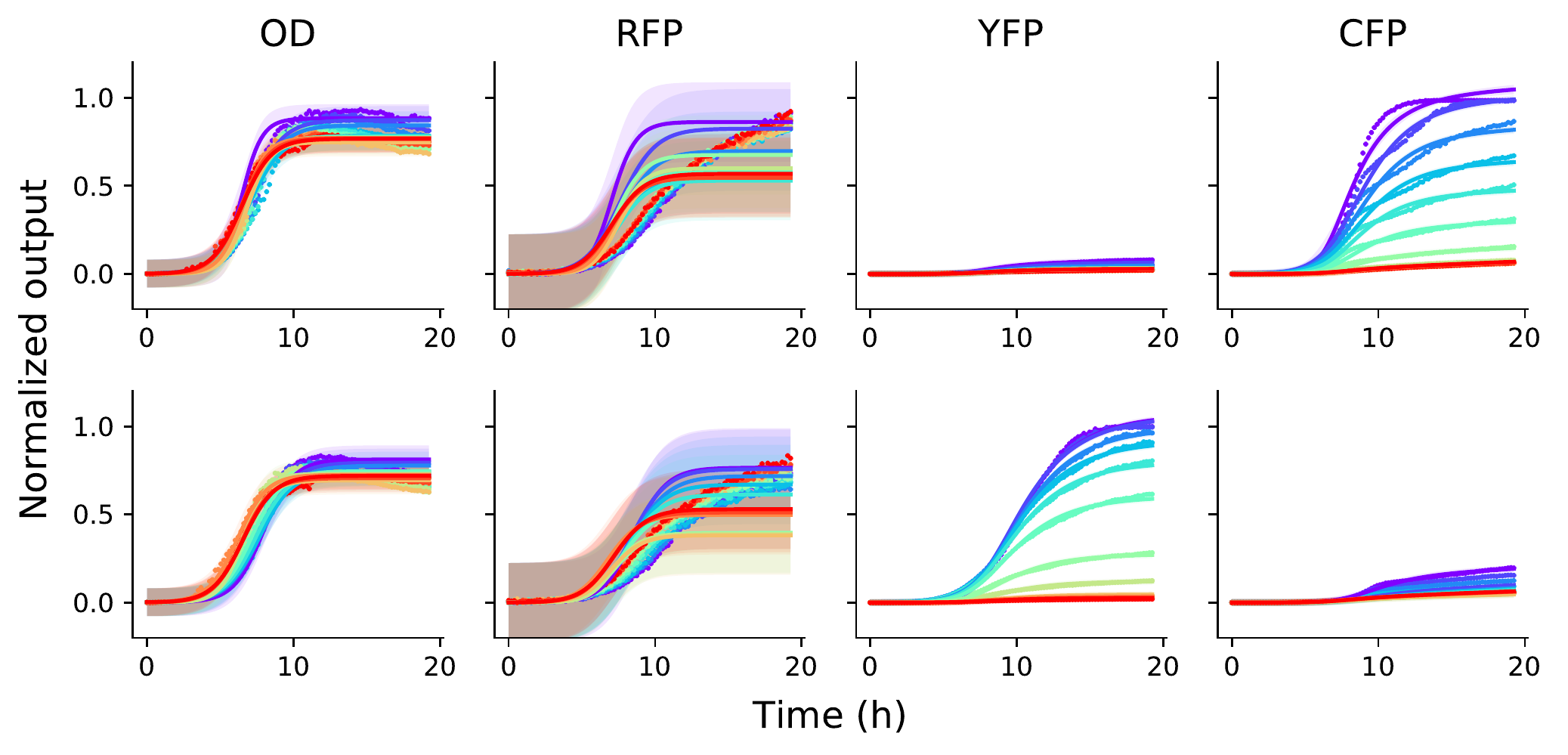}
	\caption{Summary plots for the white-box model inferred using MCMC. The posterior samples from the MCMC chain with highest log-likelihood was used to produce a posterior predictive distribution. Time-series are partitioned by treatment ($C_6$ in the top row of each panel, $C_{12}$ in the bottom row), with a color scale indicating the concentration (warm colors indicate higher concentration). Devices shown are (\textbf{A})  Pcat-Pcat and (\textbf{B}) R33-S175.}
	\label{fig:mcmc_comparisons}
\end{figure}

The solutions found by MCMC after 2 million samples were not as convincing as those found after 500 epochs of VI-HDS.
We found that the RFP signal was poorly reconstructed, with the model showing faster dynamics than the data, and the posterior predictive distribution having higher variance than for other signals (Figure~\ref{fig:mcmc_comparisons}).
With even more MCMC samples, perhaps a better parameter regime could be found.

In summary, the shortcomings of MCMC are clearly visible. 
A sequence of 1-2 million likelihood evaluations is the absolute minimum required to find a reasonable solution to the inference problem, whereas in VI-HDS, a sequence of 500 epochs is sufficient. 
As each epoch incorporates 100 importance samples, there are essentially 50,000 likelihood function evaluations, which is 40 times fewer than was required of MCMC.

\onecolumn
\section{Supplementary Figures}

~~~~~~~~~~

\begin{figure}[ht]
	\centering
	\includegraphics[width=0.7\linewidth]{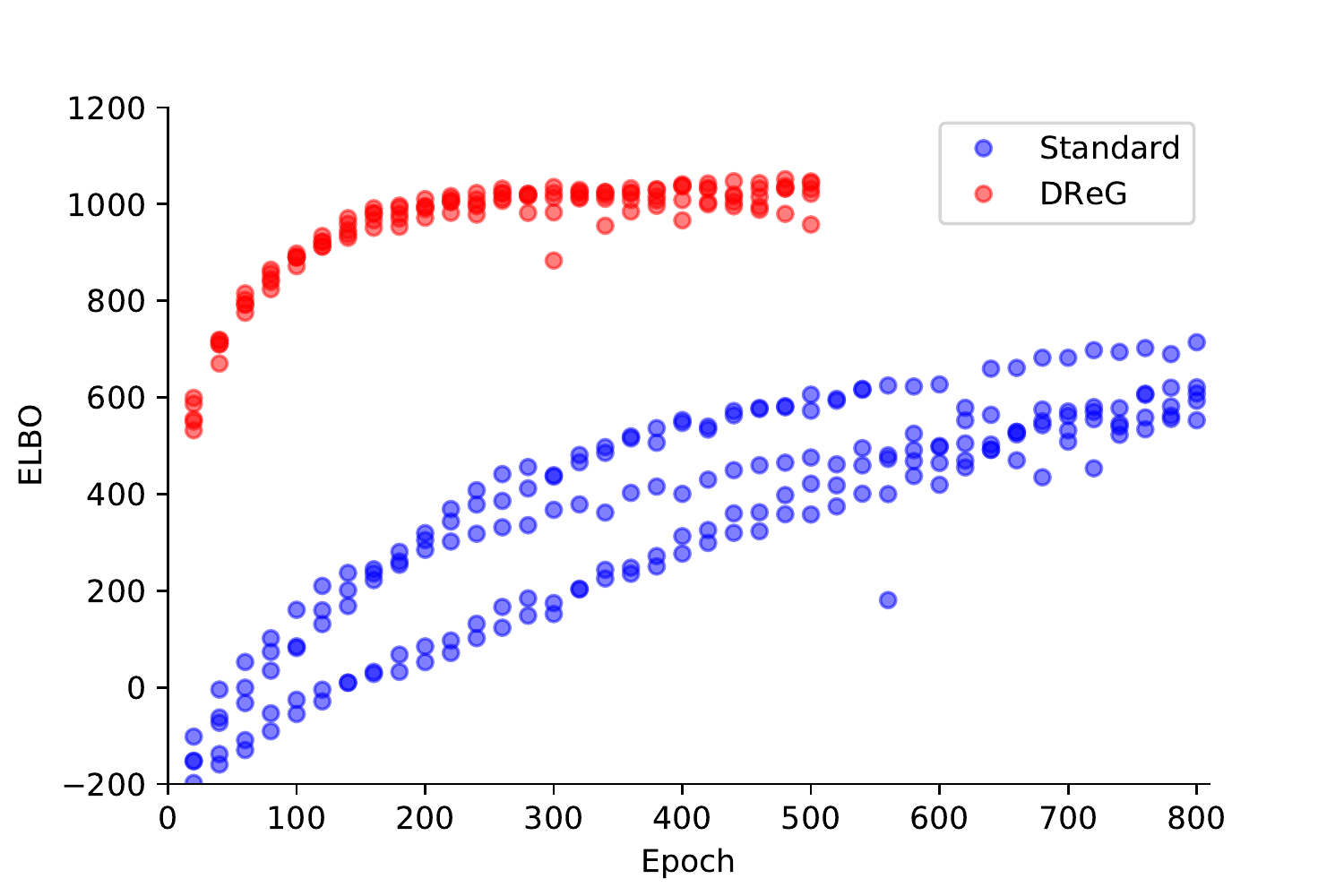}
	\caption{The convergence of the ELBO is improved by the DReG estimator. Shown are 5 independent evaluations of the VI method applied to the prescribed \emph{constant} model, using the standard gradient estimator (blue) and the DReG estimator (red). Each reported ELBO score is the average of a 4-fold cross-validation at a given epoch.}
	\label{fig:elbo}
\end{figure}

\FloatBarrier

\begin{figure}[ht]
	\centering
	\newcommand{\localwidth}{0.46\linewidth}
	\begin{tabular}{ll}
		{\sf A} & {\sf G} \\
		\includegraphics[width=\localwidth]{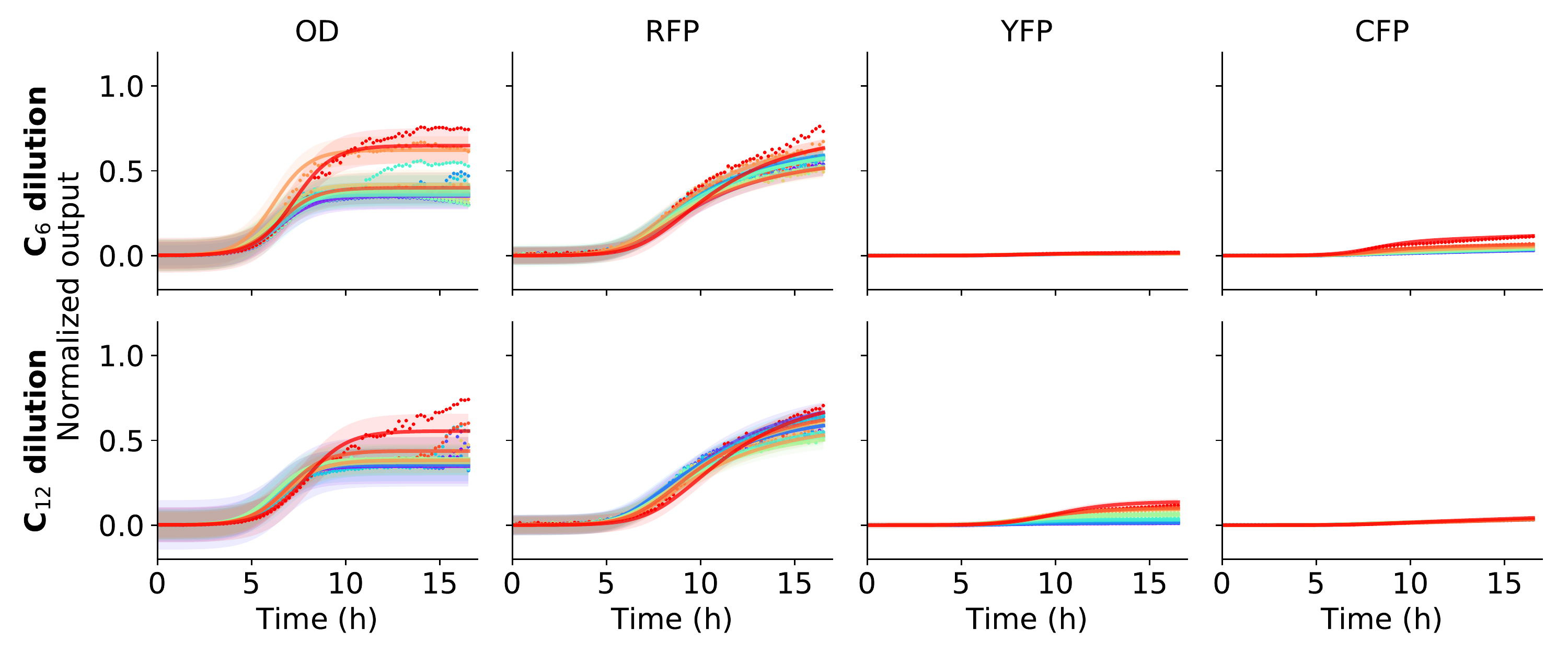} & 
		\includegraphics[width=\localwidth]{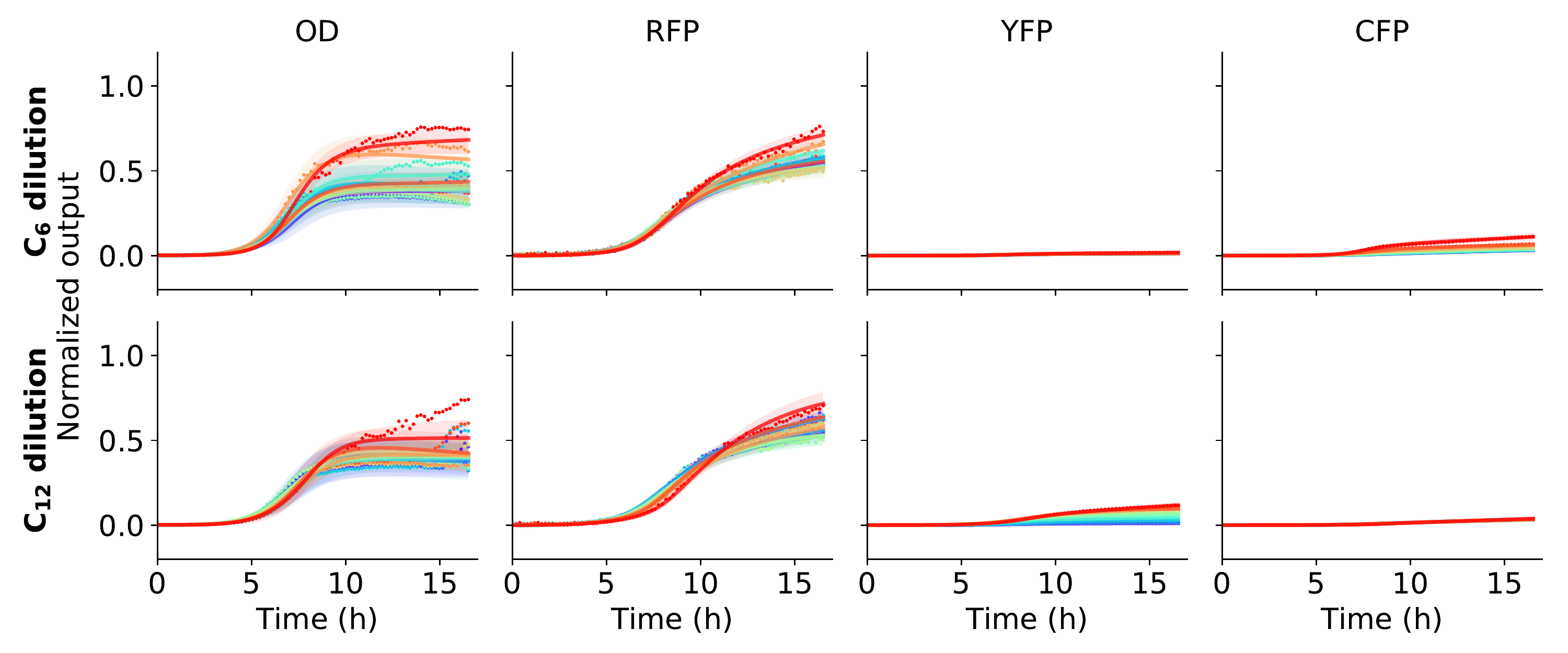} \\
		{\sf B} & {\sf H} \\
		\includegraphics[width=\localwidth,trim=0 0 0 30, clip]{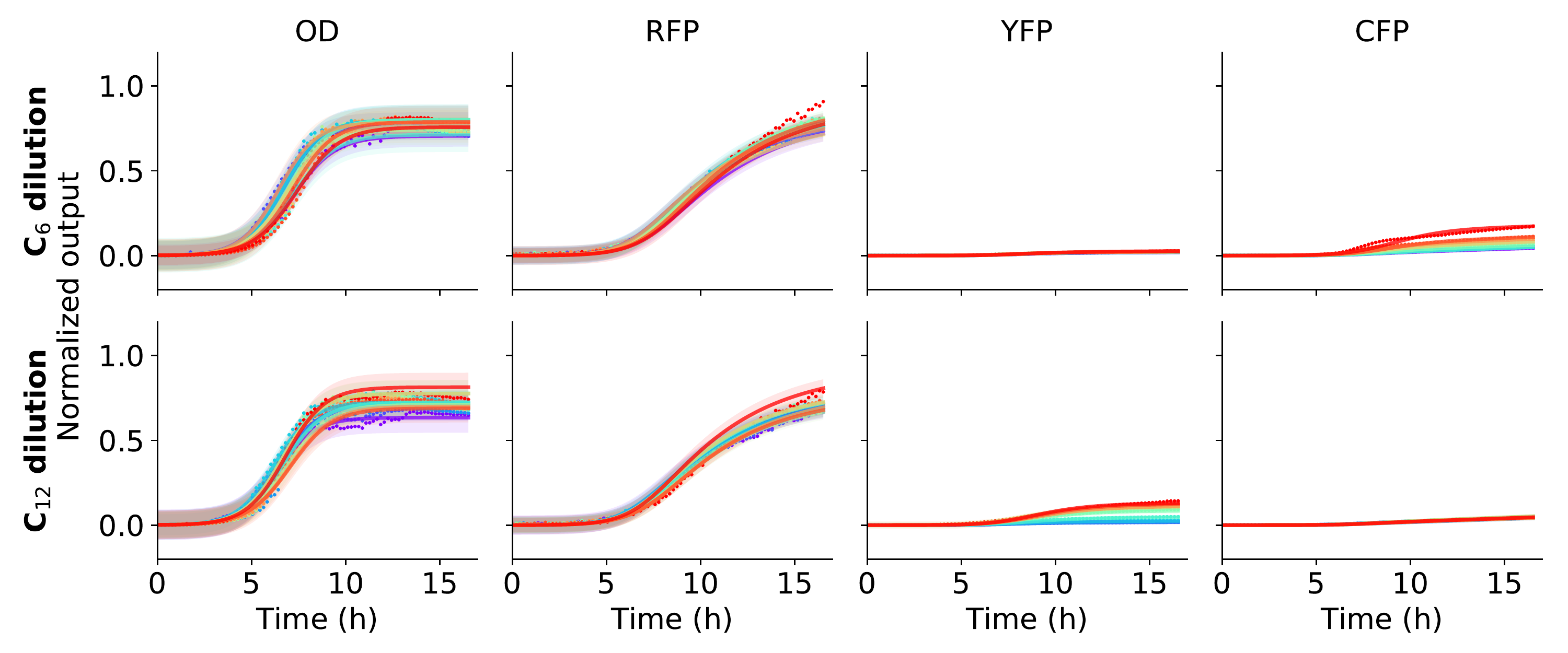} &
		\includegraphics[width=\localwidth,trim=0 0 0 30, clip]{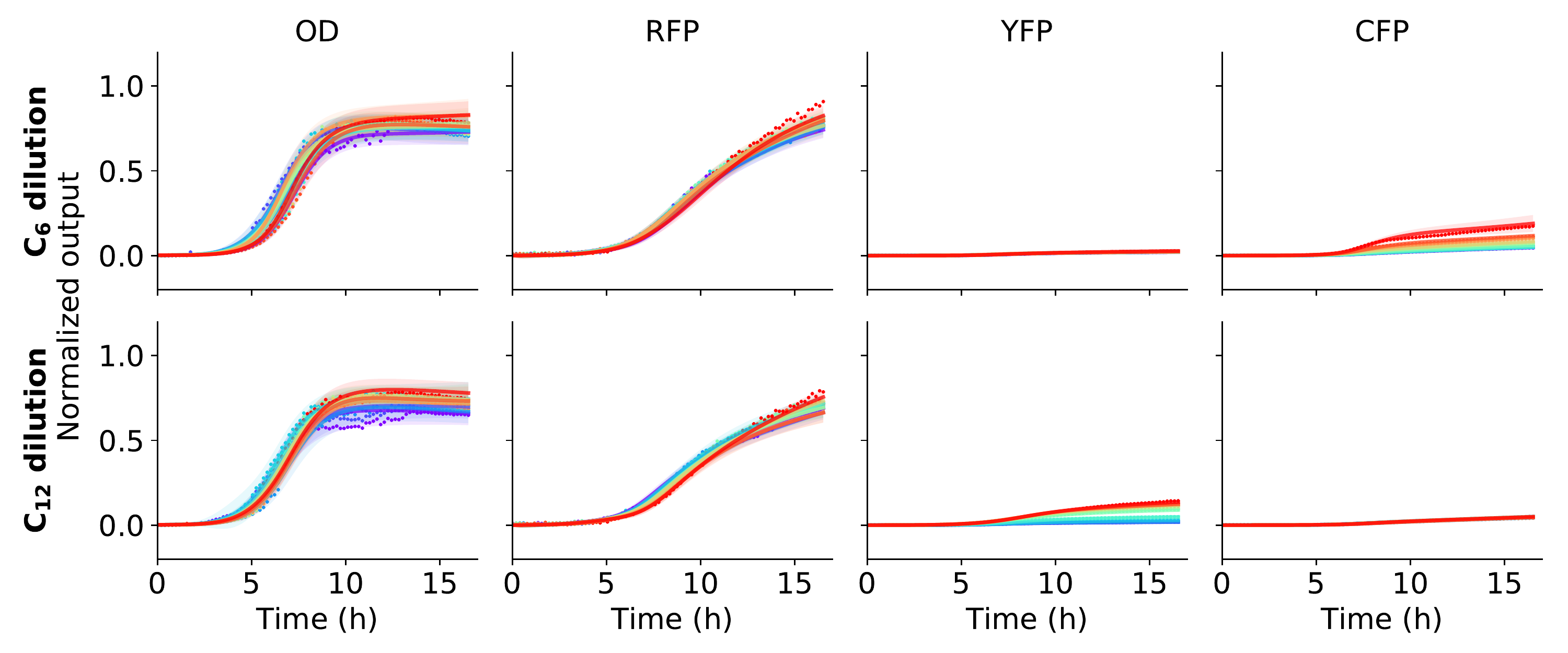} \\
		{\sf C} & {\sf I} \\
		\includegraphics[width=\localwidth,trim=0 0 0 30, clip]{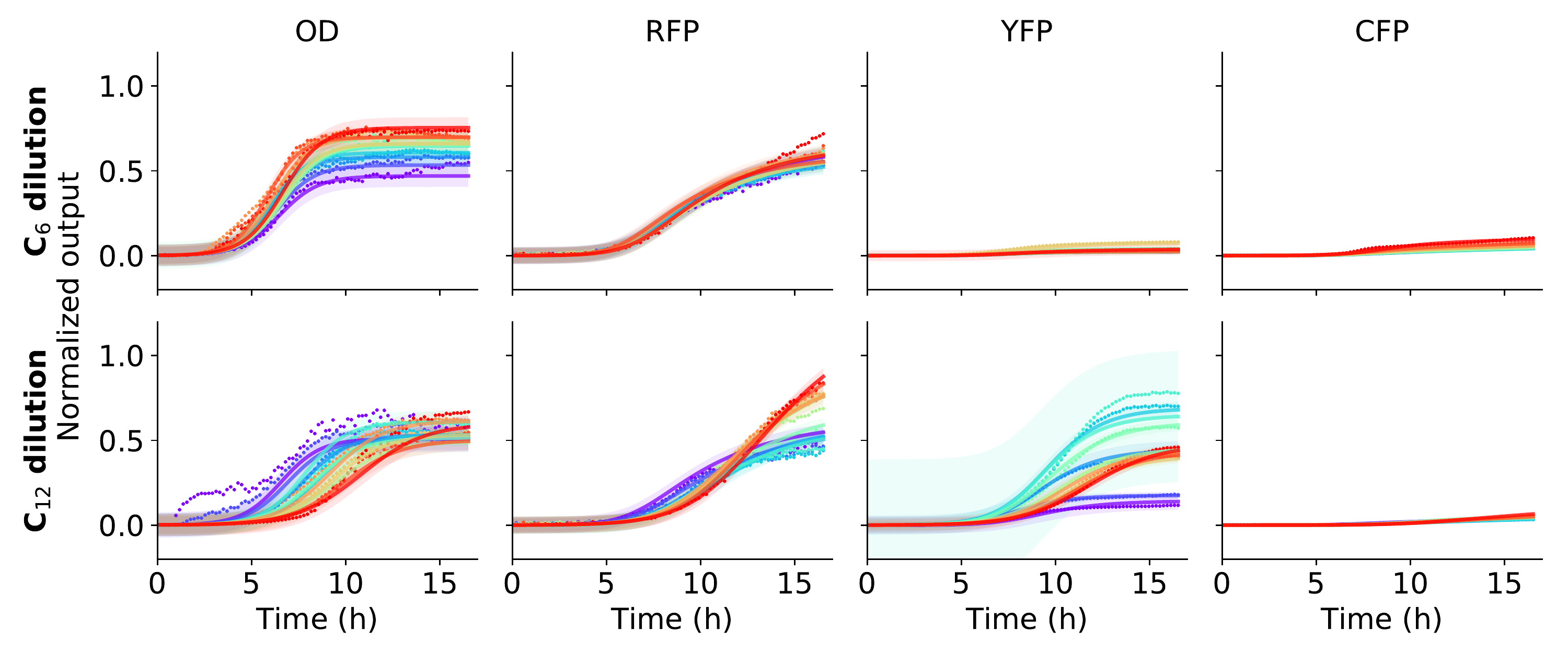} &
		\includegraphics[width=\localwidth,trim=0 0 0 30, clip]{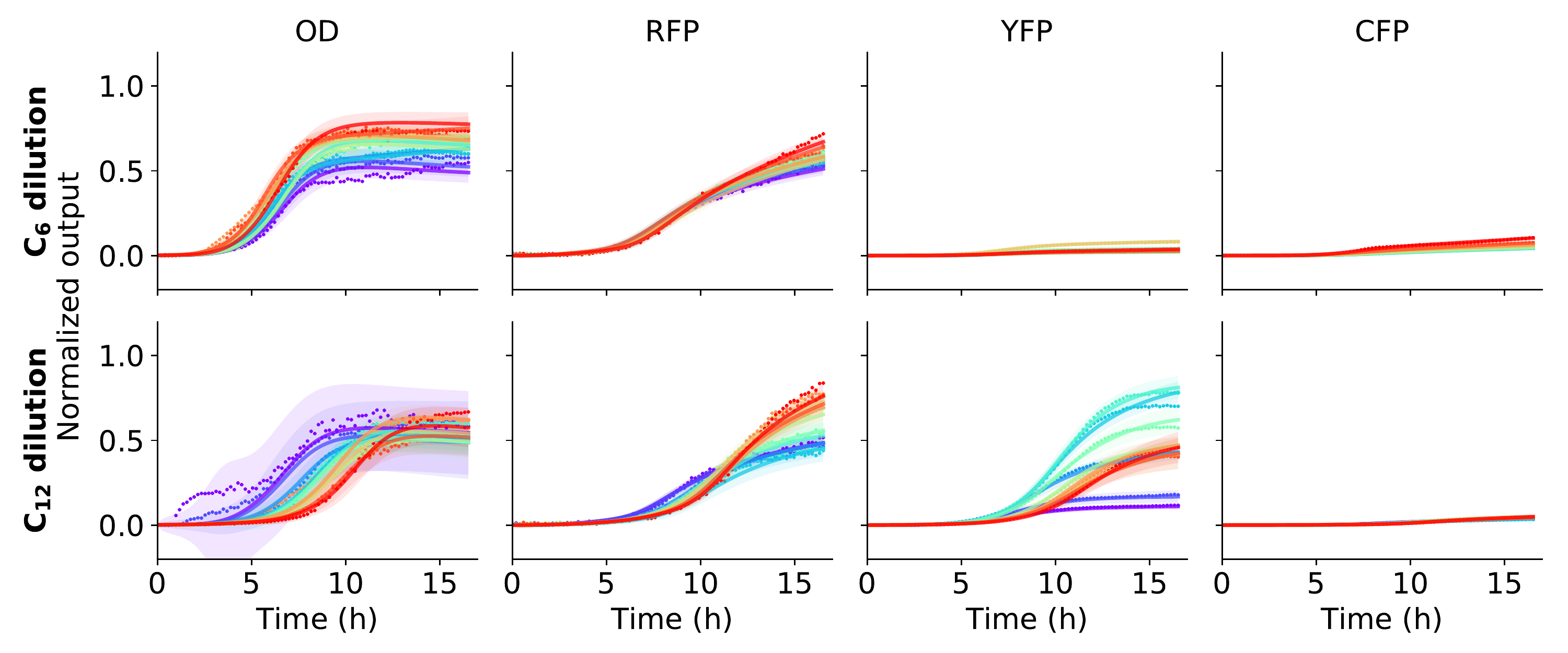} \\
		{\sf D} & {\sf J} \\
		\includegraphics[width=\localwidth,trim=0 0 0 30, clip]{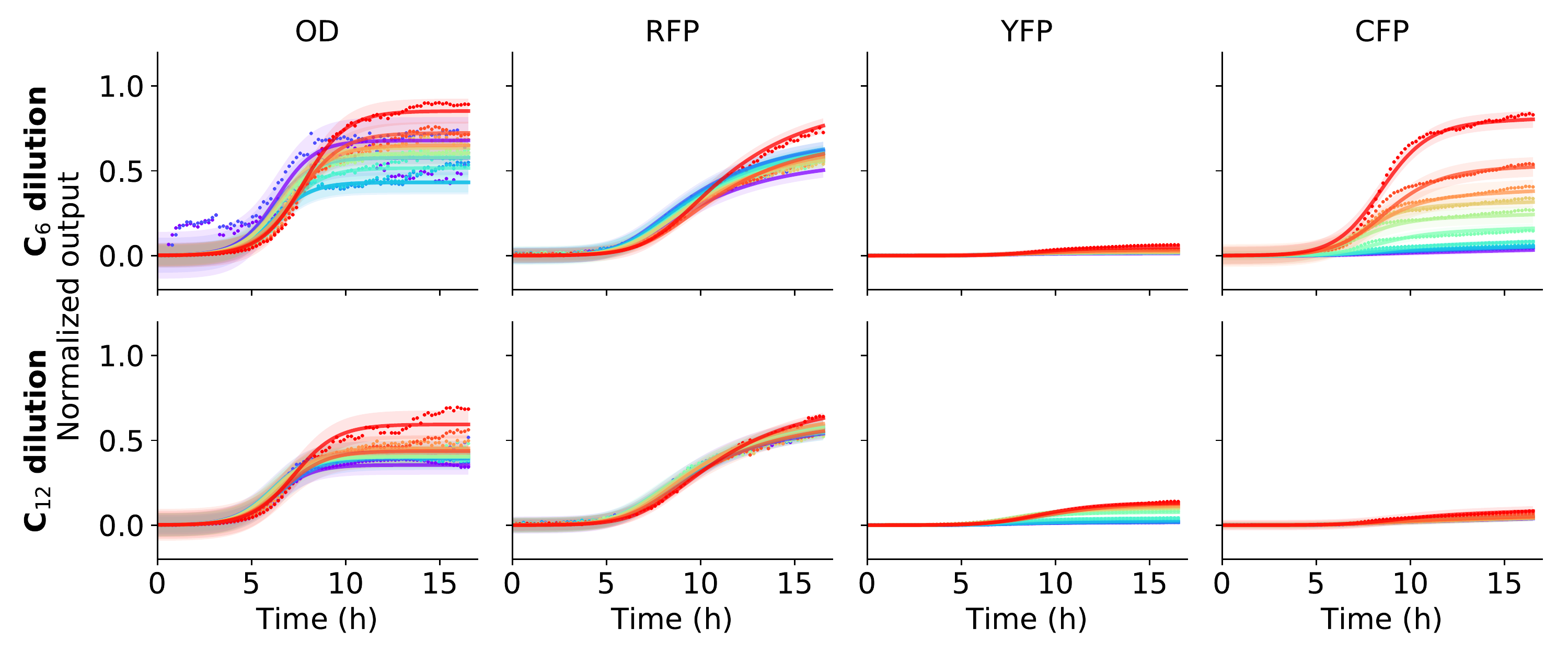} &
		\includegraphics[width=\localwidth,trim=0 0 0 30, clip]{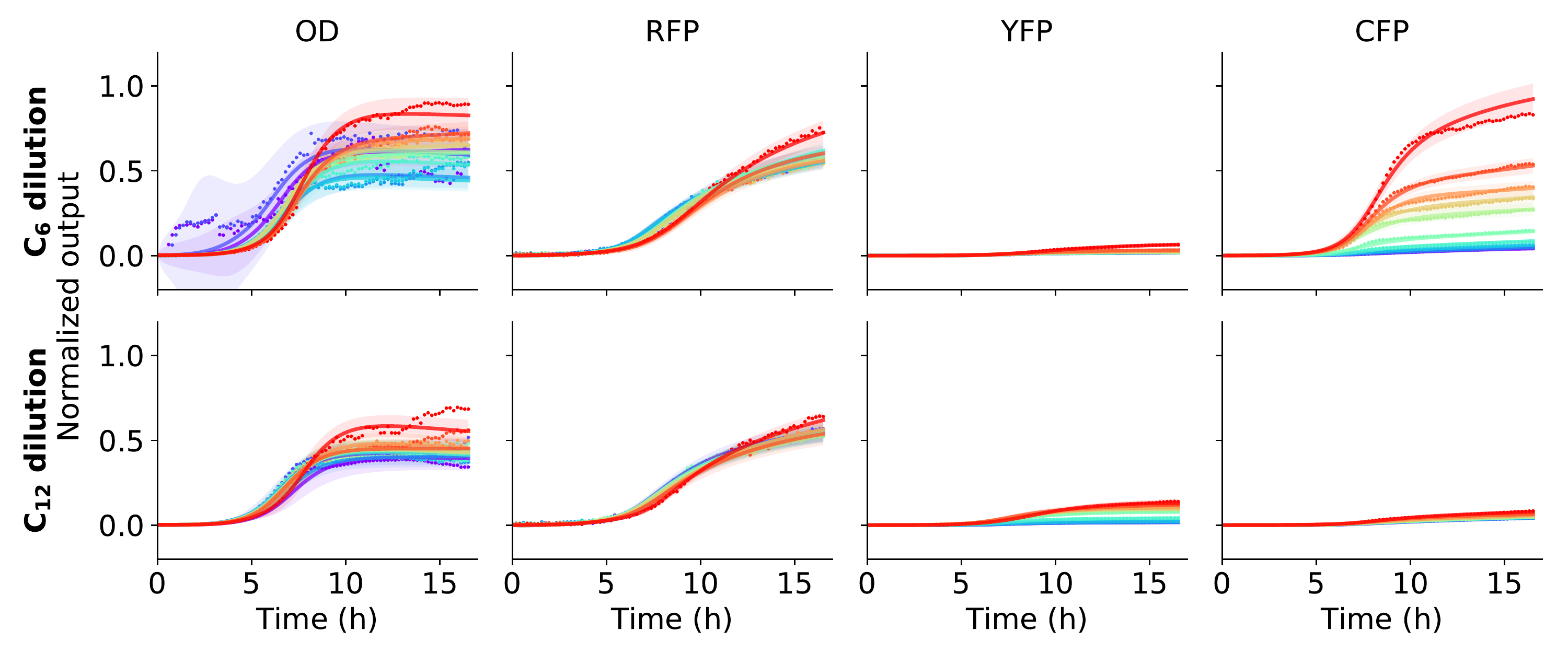} \\
		{\sf E} & {\sf K} \\
		\includegraphics[width=\localwidth,trim=0 0 0 30, clip]{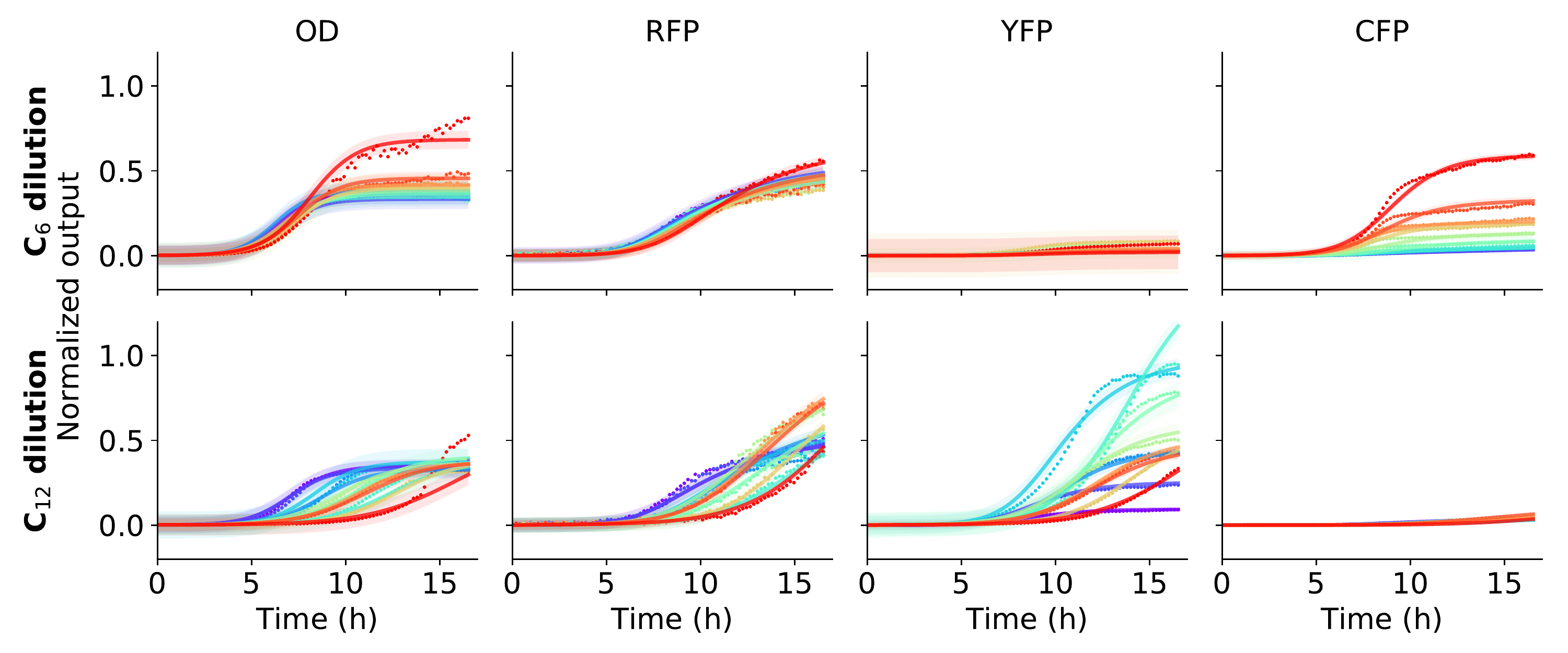} &
		\includegraphics[width=\localwidth,trim=0 0 0 30, clip]{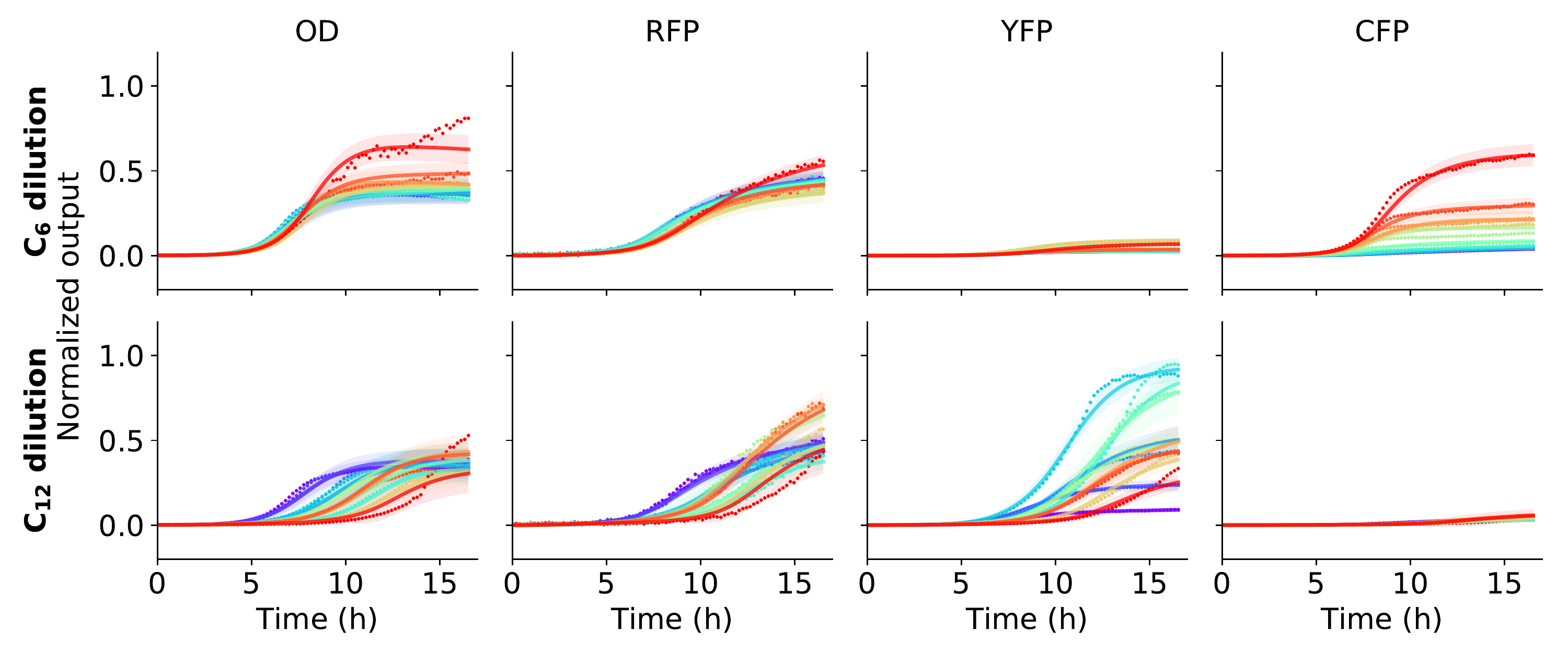} \\
		{\sf F} & {\sf L} \\
		\includegraphics[width=\localwidth,trim=0 0 0 30, clip]{figures/plot_\modelWB_R33S175_Y81C76.pdf} &
		\includegraphics[width=\localwidth,trim=0 0 0 30, clip]{figures/plot_\modelBB_R33S175_Y81C76.pdf}
	\end{tabular}
	\caption{Summary plots for the white-box model. Time-series are partitioned by device and by treatment ($C_6$ or $C_{12}$), with a color scale indicating the concentration (warm colors indicate higher concentration).
		Devices shown are (\textbf{A})  Pcat-Pcat, (\textbf{B}) RS100-S32, (\textbf{C}) RS100-S34, (\textbf{D}) R33-S32, (\textbf{E}) R33-S34 and (\textbf{F}) R33-S175.}
	\label{fig:xval}
\end{figure}

\newcommand{\individualXval}[3]{
	\begin{figure}[p]
		\centering
		\includegraphics[width=#3\columnwidth]{figures/individual_\modelWB_#1_Y81C76.png}
		\hspace{0.5cm}
		\includegraphics[width=#3\columnwidth]{figures/individual_\modelBB_#1_Y81C76.png}
		\caption{Posterior predictive distribution for 4-fold cross-validation experiment: #2 device. The left half shows the white-box model, and the right half shows the black-box model comparisons. Within each half, the left batch of four columns are $C_6$ treatments, and the right batch of four columns are $C_{12}$ treatments, as detailed to the left of each row.}
		\label{fig:individual_#1}
	\end{figure}
}

\individualXval{Pcat}{Pcat-Pcat}{0.45}
\individualXval{RS100S32}{RS100-S32}{0.45}
\individualXval{RS100S34}{RS100-S34}{0.45}
\individualXval{R33S32}{R33-S32}{0.45}
\individualXval{R33S34}{R33-S34}{0.45}
\individualXval{R33S175}{R33-S175}{0.32}

\begin{figure}[ht]
	\centering
	\includegraphics[width=0.7\linewidth]{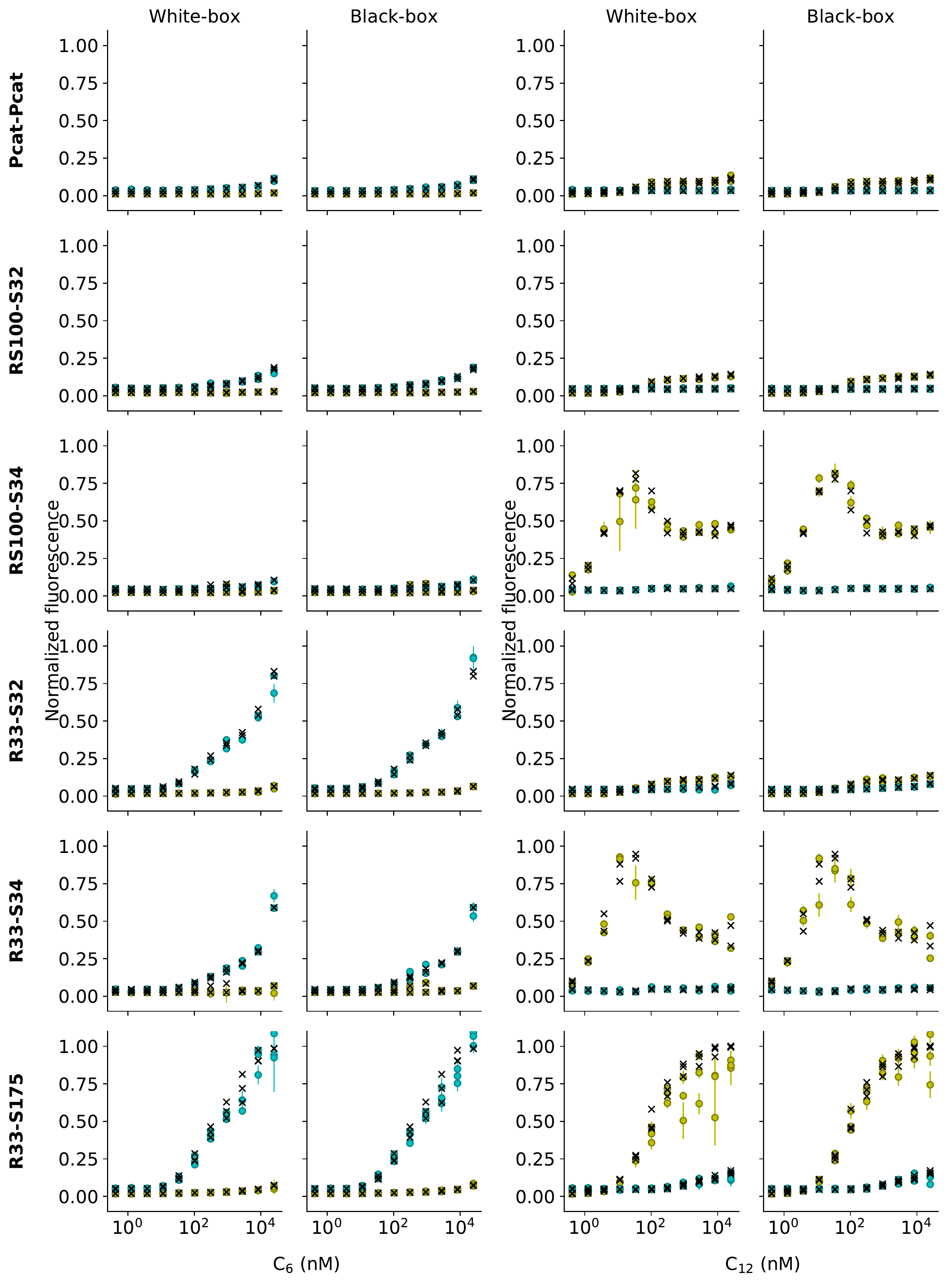}
	\caption{Treatment response plots}
	\label{fig:treatments}
\end{figure}

\begin{figure}[ht]
	\includegraphics[width=\linewidth]{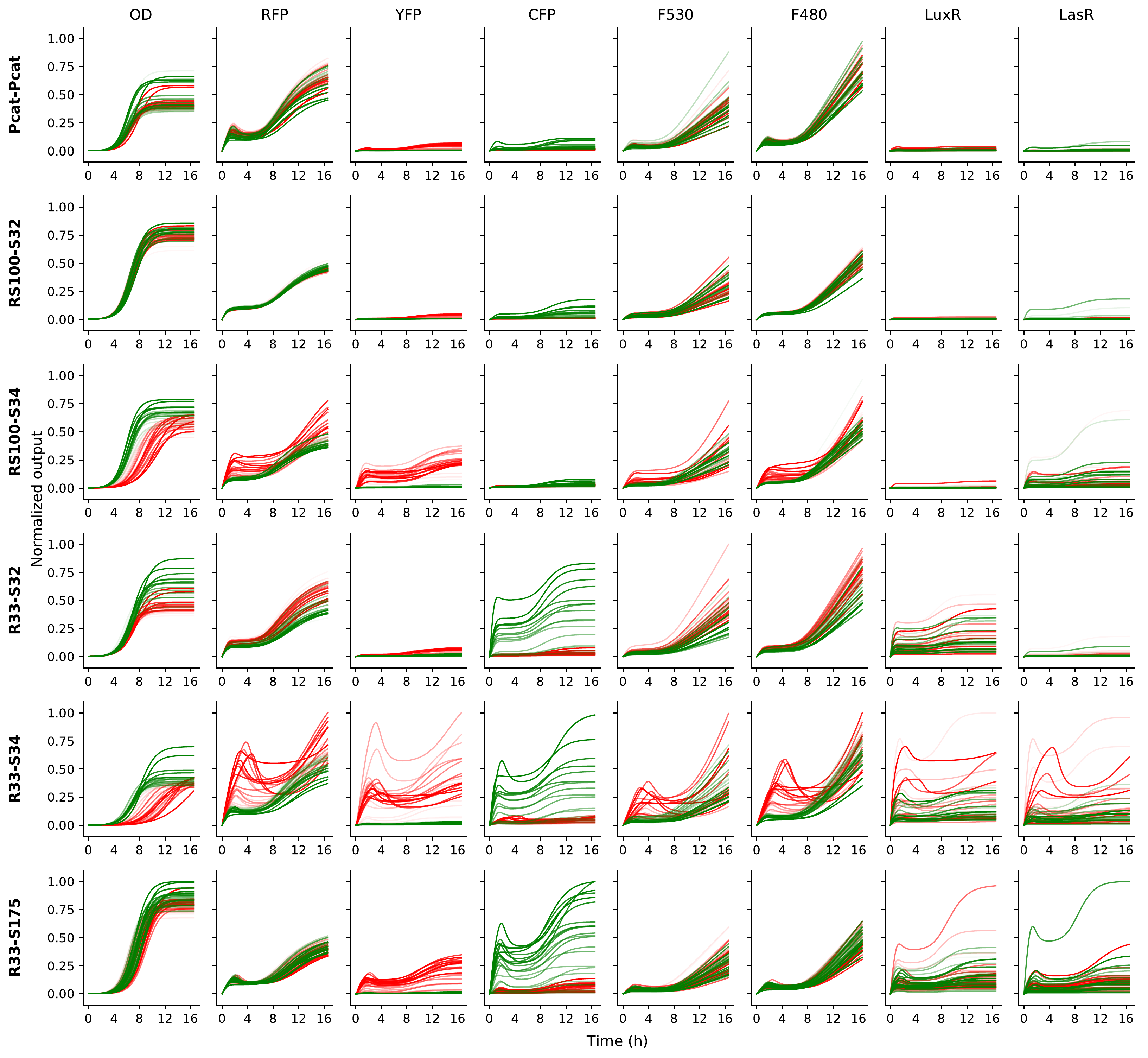}
	\caption{Predicted concentration dynamics of the hidden species ($\vx$) in the white-box model. In each panel, the traces are coloured according to whether they correspond to cultures treated with $C_6$ (green) or $C_{12}$ (red).}
	\label{fig:latentsWB}
\end{figure}

\begin{figure}[ht]
	\includegraphics[width=\linewidth]{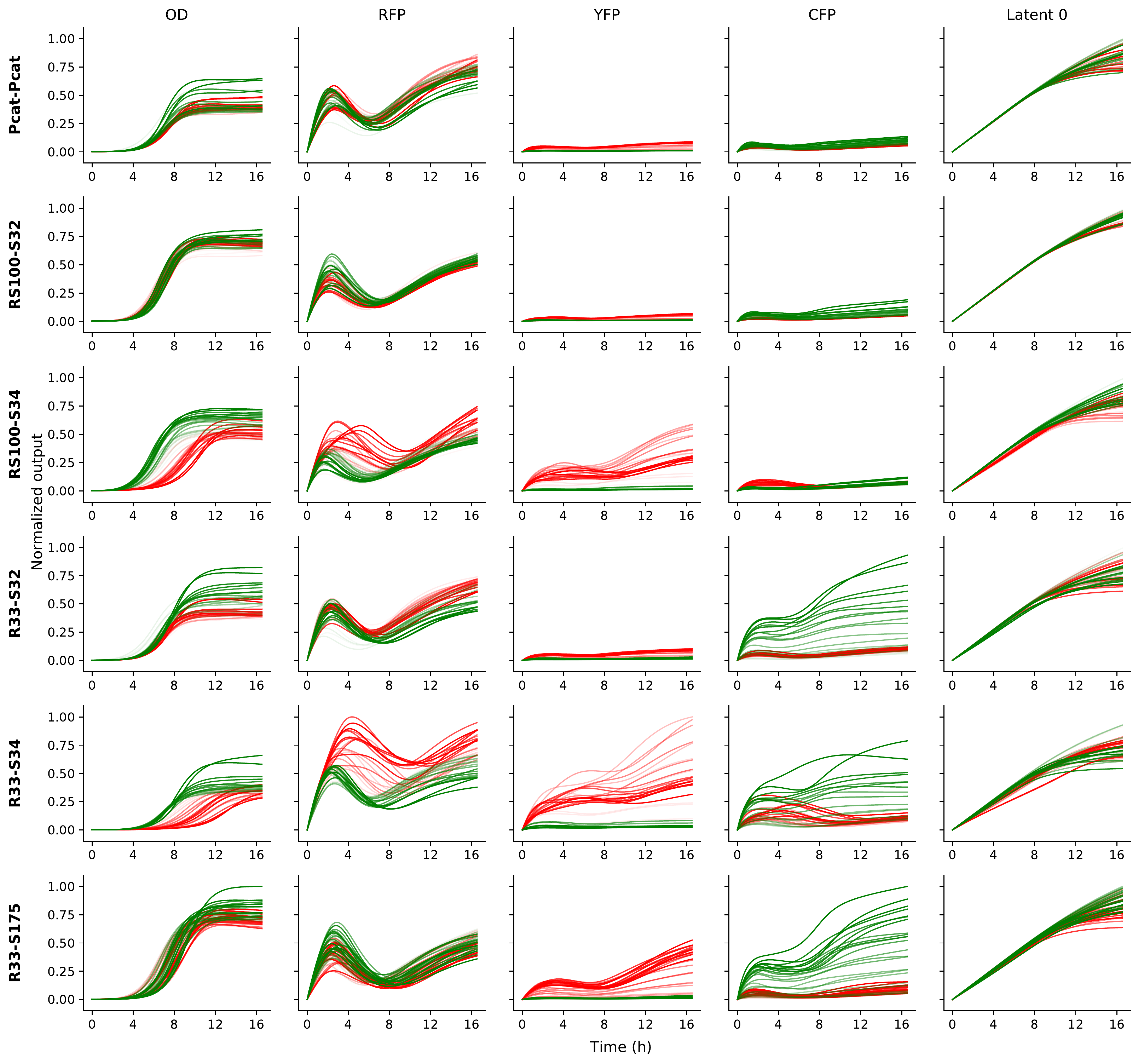}
	\caption{Predicted concentration dynamics of the hidden species ($\vx$) in the black-box model. In each panel, the traces are coloured according to whether they correspond to cultures treated with $C_6$ (green) or $C_{12}$ (red).}
	\label{fig:latentsBB}
\end{figure}

\end{document}